%% file: root_ieee_priv.tex
\documentclass[letterpaper, conference, 10pt]{ieeeconf} %{IEEEtran} %{ieeeconf}

\IEEEoverridecommandlockouts                              % This command is only needed if 
\usepackage{cite}
\usepackage[cmex10]{amsmath}
\usepackage[utf8]{inputenc}
\usepackage{algorithm}
\usepackage{algorithmic}
\makeatletter
\newcommand\fs@norules{\def\@fs@cfont{\bfseries}\let\@fs@capt\floatc@ruled
  \def\@fs@pre{}%
  \def\@fs@post{}%
  \def\@fs@mid{\kern3pt}%
  \let\@fs@iftopcapt\iftrue}
\makeatother
\floatstyle{ruled} % {norules} will delete the lines  
\restylefloat{algorithm}

\usepackage{microtype}
\usepackage{graphicx}
\usepackage{subfigure}
\usepackage{booktabs} % for professional tables

\usepackage{comment}

\usepackage{hyperref}

\usepackage{tikz}

\usepackage{amsmath}
\usepackage{amssymb,amsfonts,bm}
\usepackage{varwidth}
\usepackage{threeparttable}
\usepackage{comment}
\newcommand{\diag}{\mathop{\rm diag}}

\newcommand{\Tr}{\mathop{\rm Tr}}
\newcommand{\BBE}{\mathop{\mathbb{E}} }
\newtheorem{remark}{Remark}

\newtheorem{theorem}{Theorem}

\usetikzlibrary{arrows,positioning, decorations.pathreplacing,decorations.pathmorphing} 
\tikzset{
    >=stealth',
    punkt/.style={
           rectangle,
           rounded corners,
           draw=black, very thick,
           text width=6.5em,
           minimum height=2em,
           text centered},
    pil/.style={
           ->,
           thick,
           shorten <=2pt,
           shorten >=2pt,},
    decoration={brace},
  	tuborg/.style={decorate},
	tubnode/.style={midway, right=2pt}      
}
\pgfdeclarelayer{background}
\pgfdeclarelayer{foreground}
\pgfsetlayers{background,main,foreground}
\tikzstyle{block} = [draw, text width=4em, text centered, fill=blue!20, rectangle, 
    minimum height=3em, minimum width=4em]
\tikzstyle{sum} = [draw, fill=blue!20, circle, node distance=0.01cm]
\tikzstyle{input} = [coordinate]
\tikzstyle{output} = [coordinate]
\tikzstyle{pinstyle} = [pin edge={to-,thin,black}]
\tikzstyle{blockAtt} = [draw, fill=red!20, rectangle, 
    minimum height=3em, minimum width=4em,text width=4em, text centered]
\tikzstyle{annot} = [text width= 3em, text centered]      
\begin{document}
%
% paper title
% Titles are generally capitalized except for words such as a, an, and, as,
% at, but, by, for, in, nor, of, on, or, the, to and up, which are usually
% not capitalized unless they are the first or last word of the title.
% Linebreaks \\ can be used within to get better formatting as desired.
% Do not put math or special symbols in the title.
% \title{cloned-[CDC 18] submission - Understanding Compressive Adversarial Privacy}
\title{\LARGE \bf 
Understanding Compressive Adversarial Privacy}

\author{
Xiao Chen, Peter Kairouz, Ram Rajagopal
%, Lalitha Sankar $^{3}$
% <-this % stops a space
% \thanks{*This work was not supported by any organization}
% <-this % stops a space
\thanks{X. Chen is with Department of Civil and Environmental Engineering, Stanford University, CA 94305, USA.
        {\tt\small markcx@stanford.edu }}%
\thanks{P. Kairouz is with Google, 1600 Amphitheatre Parkway, Mountain View, CA 94043, Department of Civil and Environmental Engineering, and Department of Electrical Engineering, Stanford University,
        Stanford, CA 94305, USA.
        {\tt\small 
kairouzp@stanford.edu}}%
\thanks{R. Rajagopal is with Department of Civil and Environmental Engineering, and Department of Electrical Engineering, Stanford University,
        Stanford, CA 94305, USA.
        {\tt\small ramr@stanford.edu}}%
\thanks{This work is partially supported by NSF CAREER Award ECCS-1554178, NSF CPS Award \#1545043 and DOE SunShot Office Solar Program Award.}
}

% make the title area
\maketitle
\thispagestyle{empty}
\pagestyle{empty}

% As a general rule, do not put math, special symbols or citations
% in the abstract
\begin{abstract}
Designing a data sharing mechanism without sacrificing too much privacy can be considered as a game between data holders and malicious attackers. This paper describes a compressive adversarial privacy framework that captures the trade-off between the data privacy and utility. We characterize the optimal data releasing mechanism through convex optimization when assuming that both the data holder and attacker can only modify the data using linear transformations. We then build a more realistic data releasing mechanism that can rely on a nonlinear compression model while the attacker uses a neural network. We demonstrate in a series of empirical applications that this framework, consisting of compressive adversarial privacy, can preserve sensitive information.        
\end{abstract}
% and additive perturbations.
% no keywords

% For peer review papers, you can put extra information on the cover
% page as needed:
% \ifCLASSOPTIONpeerreview
% \begin{center} \bfseries EDICS Category: 3-BBND \end{center}
% \fi
%
% For peerreview papers, this IEEEtran command inserts a page break and
% creates the second title. It will be ignored for other modes.
\IEEEpeerreviewmaketitle

\input{priv_data_game_intro}

\input{priv_data_game_state}

\input{priv_data_game_linear}
%\input{priv_data_game_nonlinear}
\input{cdc_priv_data_game_nonLinear}
\input{priv_data_game_thm_guarantee}
\input{priv_data_game_conclusion_discuss}
\input{priv_data_g_appendix.tex}

% \vspace*{-0.19in}
\bibliography{priv_icml_ref}
\bibliographystyle{IEEEtran}
% conference papers do not normally have an appendix

% that's all folks
\end{document}

%% file: priv_data_game_intro.tex
\section{Introduction}
Machine learning has progressed dramatically in many real-life tasks such as classifying image\cite{Oshri:2018:IQA:3219819.3219924}, processing natural language\cite{sutskever2014sequence}, predicting electricity consumption\cite{sevlian2018scaling}, and many more. These tasks rely on large datasets that are usually saturated with private information. Data holders who want to apply machine learning techniques may not be cautious about what additional information the model can capture from training data, as long as the primary task can be solved by some model with high accuracy.

In this paper, we propose a privatization mechanism to avoid the potential exposure of the sensitive information while still preserving the necessary utility of the data that is going to be released. This mechanism largely leverages the concept of the game-theoretic approach by perturbing the data and retraining the model iteratively between data holder and malicious data attacker. Such a data perturbation idea is highly correlated with feature transformation and selection of the raw data input that correlated with private labels. %

Protecting privacy has been extensively explored in myriad literature. A popular procedure is to anonymize the identifiable personal information in datasets (e.g. removing name, social security number, etc.). Yet anonymization doesn't provide good immunity against correlation attacks. A previous study \cite{narayanan2008robust} was able to successfully deanonymize watch histories in the Netflix Prize, a public recommender system competition. Another study designed re-identification attacks on anonymized fMRI (functional magnetic resonance imaging) imaging datasets\cite{finn2015functional}. On the other hand, the Differential privacy (DP) \cite{dwork2008differential} has a strong standard of privacy guarantee and is applicable to many problems beyond database release \cite{dwork2014algorithmic}. This DP mechanism has been introduced in data privacy analysis in control and networks\cite{cortes2016differential,katewa2015protecting, huang2015controller, koufogiannis2017differential}. In particular, \cite{cortes2016differential} gave a thorough investigation on performing the centralized and distributed optimization under differential privacy constraints. In this line of research, \cite{huang2015controller} and \cite{koufogiannis2017differential} focused on the cases of dynamic data perturbations in control systems. \cite{katewa2015protecting} presented a noise adding mechanism to protect the differential privacy of network topology. % 

However, training machine learning models
with DP guarantees using randomized data often leads to a significantly reduced utility and comes with a tremendous hit in sample complexity\cite{duchi2013local, fienberg2010differential}. A recent work \cite{abadi2016deep} applied the DP concept on a deep neural network to demonstrate that a modest accuracy loss can be obtained at certain worst-case privacy levels. However, this was still a ``context-free'' approach that didn't leverage the full structure between the data input and output. %

To overcome the aforementioned challenges, we take a new holistic approach towards enabling private data publishing with consideration on both privacy and utility. Instead of adopting worst-case, context-free notions of data privacy (such as differential privacy), we introduce a context-aware model of privacy that allows the data holder to cleverly alter the data where it matters. %

%%% redraw the pgf/tikz picture 
% \input{priv_scheme_pgf}
\input{priv_pgf_scheme_v2}

%%%  add contribution 
%We propose a general min-max game between data holders and attackers, to seek a ``distribution free'' data releasing mechanism. We generalize it in the context of a data-driven approach and investigate a typical way of perturbing the data that is the compression. 
%Then we consider two cases of compressive adversarial privacy games under the assumption of attackers using a linear model or a neural network model. 
%% Two cases of compressive adversarial privacy games are carefully analyzed under the assumption that attackers use a linear model or a neural network model.

Our main contributions are listed as follows.
First, with the goal of having a ``distribution free'' data releasing mechanism and inspired by general min-max games, we investigate a typical way of perturbing the data that is the compression using the data-driven approach. As a second contribution, we formulate the interaction between data holders and attackers through convex optimization during the min-max game when both players apply linear models. A corresponding equilibrium can be found and used as the optimal strategy for the data holder to yield the altered data. The third contribution is that our thorough evaluations of realistic datasets demonstrate the effectiveness of our compressive adversarial privacy framework. Finally, we leverage the mutual information to validate that sensitive information can be protected from the privatized data. %

The remainder of our paper is arranged as follows. In section \ref{cdc:priv:game:state:sec:2}, we introduce the general adversarial privacy game. Section \ref{cdc:cap:statement} describes the compressive adversarial privacy game with several cases of realistic data analyses. Section \ref{cdc:quantify:priv:garentee:s4} describes the quantification of privacy. Section \ref{cdc:conclusion:s5} concludes the paper. %   
%
% 
% \section{Related Work}

%% file: priv_pgf_scheme_v2.tex
%%%%%%%%%%%%%%%%%%%%%%%%%%
% draw schematic first 
%%%%%%%%%%%%%%%%%%%%%%%%%%
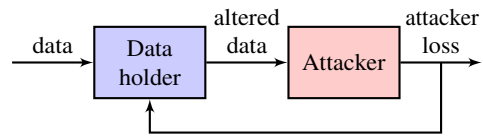
\begin{figure}[!hbpt]
\begin{center}
\begin{tikzpicture}[auto, node distance = 1.2cm, >= latex', thick, scale=0.9, every node/.style={transform shape}]
% set a bounding box 
% \path[use as bounding box] (-1, 0) rectangle (10, -2);
\node [input, name=input]{}; 
\node [block, right= of input] (DataHolder) {Data holder};
\node [blockAtt, right = of DataHolder] (Attacker) {Attacker};
\node [output, right = of Attacker] (outputLoss) {}; 
\draw [draw, ->] (input) -- node[annot] {data} (DataHolder);
\draw [draw, ->] (DataHolder) -- node[annot] {altered data} (Attacker);
\draw [draw, ->] (Attacker) -- node [annot, name=attLoss] {attacker loss} (outputLoss);
\path [draw, ->] (attLoss) -- +(0, -1.5) -| (DataHolder); % |-  (DataHolder); 
\end{tikzpicture}
\end{center}
\caption{Data releasing schematic}
\end{figure}

%% file: priv_data_game_state.tex
\section{Privacy Preserved Releasing}
\label{cdc:priv:game:state:sec:2}
We propose a general data publishing framework by incorporating the following game concept. In general there are two roles in this data releasing game: a data holder and a data consumer. Among data consumers, some people are good users who explore the pattern and extract the value of the data. We merge the good data consumers together with data holders and address their roles from the data holder perspective, since good data user are irrelevant in this game. 
Yet there are people who also try to learn personal sensitive information from the data on purpose. We define those malicious users as attackers. We focus on the data holders and attackers for the following description of the game. %

Consider a dataset $\mathcal{D}$ which contains both the original data $X$ and the associated customers' sensitive demographic information $Y$ (e.g. account holder age, house square-footage, gender, etc.). Thus a sample $i$ has a record $(x_i, y_i) \in \mathcal{D}$. We denote the function $g$ as a general mechanism for a data holder to release the data. The released data are denoted as $\tilde{x_i}$ for the customer $i$, which can also be described as $\tilde{x_i}= g(x_i,y_i)$. Notice we don't release $y_i$ to the public because it's private information. Generally speaking, $\tilde{X}=g(X, Y)$. Let the function $h$ represent the adversarial hypothesis, e.g. the estimated outcome $\hat{Y}=h(\tilde{X})$. The attacker would like to minimize the inference loss on private labels, namely $\ell(\hat{Y}, Y)$ given some loss function $\ell$, %individual $y_i$ given $(\tilde{x_i}, {y_i})$. 
while the data holder would like to maximize the attacker's loss, and in principle, also wants to preserve the quality of the released data for research purposes. This data quality is characterized by some distance function measuring between the original data and the altered data. Therefore, we formulate a min-max game between the data holder and the attacker as follows: 
\begin{align}
\max_{g \in \mathcal{G}}\Big\{\min_{h \in \mathcal{H}} \ell\Big(h\big(g(X, Y)\big), Y\Big)   \Big\} \label{cdc:prob:formulation:general1}\\
s.t. \quad  d\big( X, g(X, Y) \big) \leq \gamma \label{cdc:prob:formulation:general2},
\end{align}
where $d()$ could be some distance function, such as Total Variation (TV), Wasserstein-1, or Frobenius norm, etc. \cite{nguyen2009surrogate, arjovsky2017wasserstein}, and $\gamma$ is a hyper-parameter. The constraint ensures that the released data will not be distorted too much from the original data. % 

This framework allows an attacker to incorporate any loss functions and design various adversarial inference models, which typically take the released data to predict the personal information. Given such a challenge, the data publisher has to design a good privatization mechanism $g$ and $\gamma$ to deteriorate the attacker's performance, which are also data dependent. For simplicity, we focus on the supervised learning setting in this work, but the concept can potentially be extended to the unsupervised learning. %We demonstrate that, in some cases, the data publisher can fulfill this goal.      
%

%
%

%
%%%%%%%%%%%%%%%%%%%%%%%%%%%
% linear compression on X %
%%%%%%%%%%%%%%%%%%%%%%%%%%%
%

%% file: priv_data_game_linear.tex
%%%%%%%%%%%%%%%%%%%%
\section{Compressive Adversarial Privacy}
\label{cdc:cap:statement}
%%%%%%%%%%%%%%%%%%%% 
A typical method to enforce data privacy is data compression. This method is well studied in \cite{zhou2009differential} from a theoretical point of differential privacy. In reality, data compression is used in many applications such as text messaging and video transmission to protect the privacy. In this section, we extend the general min-max framework to a compression approach, namely a compressive adversarial privacy framework, as shown in Figure \ref{cdc:fig:compress:diag}\\  
\input{priv_pgf_autoencoder.tex} We focus on two scenarios to illustrate concrete privatization mechanisms. The first one is the linear compression when an attacker takes a linear model. The second one is the non-linear compression when an attacker uses a neural network. We evaluate of both cases based on real data.  
% \textcolor{red}{TODO: adding topic sentences to give a overview of the following content plus some context examples}
\subsection{Linear compression with continuous label}
We introduce the case of an attacker who uses a linear model and the least squared loss function to infer private information through released data. This case is practical especially when private labels are continuous and have a linear relationship with the original data. We denote the original data matrix $X$ and altered data matirx $\tilde{X} \in \mathbb{R}^{n\times p}$, where $n$ is number of samples, $p$ is number of features, and $\mathbb{R}$ is the set of real numbers. Such a data matrix contains individual samples $x_i$ and $\tilde{x}_i \in \mathbb{R}^p$ respectively, where $i=1,\dots, n$. The private-info matrix $Y \in \mathbb{R}^{n\times d}$ consists of $y_i \in \mathbb{R}^d$ that each sample $i$ has $d$ types of private labels. %

Consider a data holder who has a simple data-releasing mechanism that applies a linear transformation of $X$, namely projecting it down to lower dimensions to protect confidential information, i.e. $Z=XA$, where the matrix $A \in \mathbb{R}^{p\times k}, k < p$. Hence, $Z \in \mathbb{R}^{n\times k}$. In order to release meaningful data that still can be utilized by a majority of good users, the data holder performs a linear operation by multiplying $B \in \mathbb{R}^{k\times p}$ on $Z$ and recovers it back to the same dimension as $X$, yielding $\tilde{X} = XAB$. The attacker fits a linear model to minimize the mean squared loss that is $\frac{1}{n} \sum_{i=1}^n \|\tilde{\Theta}^T\tilde{x}_i - y_i\|_2^2 = \frac{1}{n} \sum_{i=1}^n \| \tilde{\Theta}^T B^TA^T x_i - y_i\|_2^2$, where $\tilde{\Theta} \in \mathbb{R}^{p \times d}$. Because the domain of $\tilde{\Theta}^TB^T$ is contained in $\Theta^T$ which is in $\mathbb{R}^{d\times k}$. This attacker's loss is lower bounded by  
\begin{align}
 \min_{\Theta}\frac{1}{n}\sum_{i=1}^n\|\Theta^T A^Tx_i - y_i \|_2^2 ,
\end{align}
where $\Theta \in \mathbb{R}^{k \times d}$. 
%It's due to that $\Theta^T$ lives in $R^{d\times k}$ which includes $\tilde{\Theta}^TB^T$. 
Therefore, when the data holder maximizes the attacker's loss, we can maximize this lower bound that automatically maximizes the minimum loss of the attacker. The resulting min-max problem can be formulated as 
\begin{align}
\max_{A, B} \min_{\Theta}\frac{1}{n}\sum_{i=1}^n\|\Theta^T A^Tx_i - y_i \|_2^2 \\
s.t. \quad \|XAB - X\|_{F}^2 \leq \gamma ,   
\end{align}
where $\| \cdot \|_{F}$ is the Frobenius norm. Given $A$, we further simplify the expression by finding the best recovering matrix $\hat{B}$ in place of $B$ as follows (see \ref{priv:icml2018:appendix:recover:mat:B} for details): 
\begin{align}
\hat{B} = (A^TX^TXA)^{-1} A^TX^TX = (A^TA)^{-1}A^T = A^{\dagger} .
\label{linear:sudo:inverseA}
\end{align}
We denote the $A^{\dagger}$ to be the pseudo-inverse of $A$. The best predictor $\Theta$ for the attacker can be expressed as $\Theta = (A^T C_{xx} A)^{-1} A^T C_{xy}$, where $C_{xx}=\frac{1}{n}\sum_{i=1}^n x_ix_i^T, C_{xy}=\frac{1}{n} \sum_{i=1}^n x_iy_i^T $. Substituting $\Theta$ and $B$, we have the following problem: 
\begin{align}
 \max_{A} \bigg\{ - \Tr\Big( C_{xy}^T A(A^T C_{xx} A)^{-1}A^TC_{xy} \Big) \bigg\}\\
s.t. \quad \|X A(A^TX^TXA)^{-1}A^T X^T X - X\|_{F}^2 \leq \gamma . 
\end{align}
Notice that $C_{xx} = \frac{1}{n}X^TX$. By flipping the sign of the maximization and denoting $M = A(A^TX^TXA)^{-1}A^T$ which is a positive semidefinite matrix (see the appendix \ref{pes:priv:app:linear:tran:M:psd}), we have the following problem: 
\begin{align}
& \min_{M} \frac{1}{n}\Tr\big( C_{xy}^TXMX^TC_{xy} \big) \\
& s.t.  \qquad M \succeq 0 \\
& \qquad \|XMX^TX - X\|_{F}^2 \leq \gamma \\
& \qquad rank(M) = k . \label{prob:formulate:psd:rank:constraint} 
\end{align}
We put a rank constraint (\ref{prob:formulate:psd:rank:constraint}) because the dimension $A$ is $p\times k, (k<p)$. This problem can be further relaxed to a convex optimization by regularizing the nuclear norm of matrix $M$ as follows:
\begin{align}
\min_{M} & \frac{1}{n}\Tr\big( C_{xy}^TXMX^TC_{xy} \big) + \beta \|{M}\|_{*}  
\label{pes:linear:filter:X:param:M:cvx_form_1} \\
s.t. & \quad \|XMX^TX - X\|_{F}^2 \leq \gamma \\ 
& \quad {M} \succeq 0 , \label{pes:linear:filter:X:param:M:cvx_form_2}
\end{align}
where $\|\cdot \|_{*}$ is the nuclear norm of a matrix that heuristically controls the rank of a matrix. Such a convex relaxation allows the data publisher to find an optimal solution of $M$, and correspondingly yields the appropriate $\tilde{X}$ (see appendix \ref{priv:icml2018:cast:linear:implicit:solve}). Thus, both players can achieve an equilibrium in this game. To ensure the problem is feasible, one caveat is that we cannot pick arbitrarily small $\gamma$ without considering the aforementioned rank $k$. We note that $\tilde{X} = XAA^{\dagger}$ is a low rank-$k$ approximation of the original data matrix $X$. 
\begin{theorem}\label{cdc:svd:err:th1}
Suppose a rank-$p$ matrix $X$ consists of the singular values $\lambda_1, \dots, \lambda_p$. With the best rank-$k$ approximation $\tilde{X}_k$ under Frobenius norm, the distortion threshold $\gamma$ is at least $\sum_{i=k+1}^p \lambda_i^2$.
\end{theorem} %

We put the proof in appendix \ref{cdc:appx:proof:Th1}. The theorem reveals the relationship between setting the distortion tolerance $\gamma$ and the rank $k$. 
% \begin{proof}
% Given $X = \sum_{i=1}^p \sigma_i u_i v_i^T$, 
% the best rank-$k$ approximation $\tilde{X}_k = \sum_{i=1}^k \sigma_i u_i v_i^T$ is achieved by SVD in Frobenius norm by Eckart-Young theorem\cite{golub1987generalization}. Then $\|X-\tilde{X}_k\|_F^2 = \Tr\big( (\sum_{i=k+1}^p\sigma_i u_i v_i^T) (\sum_{i=k+1}^p\sigma_i u_i v_i^T)^T \big) = \Tr(\sum_{i=k+1}^p\sigma_i^2 ) = \sum_{i=k+1}^p \sigma_i^2$
% \end{proof}
Hence, a simple algorithm ({Algorithm \ref{pes:priv:alg:linear:gen:X_tilde}}) is proposed for the data holder to generate $\tilde{X}$.  
\begin{algorithm}[h!]
\caption{Generating $\tilde{X}$ (Linear attacker)}
\label{pes:priv:alg:linear:gen:X_tilde}
\begin{algorithmic}[1]
\STATE \textit{Input:} dataset $ (X, Y) \in \mathcal{D}$,  parameter $\gamma$, $\beta_0$, $k$, $\eta$. 
\STATE \textit{Output:} $\tilde{X}$ 
% \State incrementally control the hyper-parameter $\gamma$ and $\beta$ to check if the rank of $\tilde{M}$ equals $k$. 
\STATE partition dataset into several batches $(X,Y)$. 
\FOR {a batch of $(X,Y) \in \mathcal{D}$}
	\STATE $\hat{k}=0$, $t=0$
	\WHILE {$\hat{k} \neq k$}
    \STATE $ {M}_{t} = \arg\min_{ {M }}\{ \frac{1}{n} \Tr( C_{xy}^TX {M}X^TC_{xy}) + \beta_t \|{M}\|_{*}$ \\
    s.t. $\quad \|XMX^TX - X\|_F^2 \leq \gamma \}$  (solving the optimization in equation (\ref{pes:linear:filter:X:param:M:cvx_form_1} - \ref{pes:linear:filter:X:param:M:cvx_form_2}) with certain values of $\gamma, \beta_t, k$).  
%%% generate svd %%%    
\STATE $U_{t}, \Lambda_{t} \gets SVD({M}_{t})$ (applying Singular Value Decomposition on ${M}_t = U_t\Lambda_t U_t^T$ to get matrices $U_t$ and $\Lambda_t$.)    
\STATE $ \hat{k} = \tilde{rank}(\Lambda_t)$. check the rank of the matrix $\Lambda$ with non trivial eigenvalues. (e.g $\lambda_j > \eta \lambda_{max}, \text{where } \forall j =1,\hdots,n; \eta = 0.01$.) 
  \IF {$\hat{k} = k $} 
  \STATE \textbf{break}
  \ELSIF {$\hat{k} > k $}
  \STATE $\beta_{t+1} \gets \beta_t + \frac{\beta_t}{2}$
  \ELSIF {$\hat{k} < k $} 
  \STATE $\beta_{t+1} \gets \beta_t - \frac{\beta_t}{4} $ 
  \ENDIF 
% belong to $\{0, 1\}$. 
% If not, decompose the diagonal matrix $\Lambda = \sqrt{\Lambda} I \sqrt{\Lambda} $, then multiply it to $U$ to get $\tilde{U} = U\sqrt{\Lambda}$
% \State pick a diagonal matrix $D$ and orthonormal matrix $V$, say both are identity matrix.  
% \State $\tilde{X} \gets \tilde{U} \begin{bmatrix}
% D \\
% 0 \\
% \end{bmatrix} V^T $
  \STATE {$t = t+1$}
\ENDWHILE
\STATE $\tilde{X} \gets X{M}X^TX $
\ENDFOR
\end{algorithmic}
\vspace*{-0.04in}
\end{algorithm}

%%%
\begin{remark}
This approach can be interpreted as releasing a low dimensional approximation to a set of data, incorporating the relation between the original data and private labels, while still maintaining a certain distortion between the released data and the original data.
\end{remark}
We also discovered that a similar scheme can be applied on compressing original data with additive Gaussian noise. 
See Appendix \ref{icml:appendix:linear:compress:noised:data} for details.
%%%%
%
%%%%%%%%%%%%%%%%%%%%%%%%
%%%%%%%%%%%%%%%%%%%%%%%%
%%%%%%%%%%%%%%%%%%%%%%%%
%%%%%%%%%%%%%%%%%%%%%%%%
%%%%%%%%%%%%%%%%%%%%%%%%
%%%%%%%%%%%%%%%%%%%%%%%%
%%%%%%%%%%%%%%%%%%%%%%%%
%%%%%%%%%%%%%%%%%%%%%%%%
%%%%%%%%%%%%%%%%%%%%%%%%
%%%%%%% Results %%%%%%%%
%
% illustrating some results 
%%%%

\subsection{Case study: Power consumption data}
% \vspace{-0.25in}
The first experiment of our analysis uses the CER dataset, which was collected during a smart metering trial conducted in Ireland by the Irish Commission for Energy Regulation (CER) \cite{CER2009Dataset}. The dataset contains measurements of electricity consumption gathered from over 4000 households every 30 minutes between July 2009 and December 2010. Each participating household was asked to fill out a questionnaire about the households' socio-economic status, appliances stock, properties of the dwelling, etc.\cite{beckel2014revealing}. To demonstrate our concepts, we sampled a portion of the customers who has valid entries of demographic information, e.g. number of appliances and floor area of the individual house. In the following experiment, we treat floor area as private data $Y$. %

Throughout the case simulations, we extract the four-week time series in September 2010. Since the power consumption (in kilowatts) is recorded every 30 minutes, there are $2 \times 24 \times 28 = 1344$ entries for a single household. To simplify the input dimension and avoid the over-fitting issue from raw input, we compute a set of features on the electricity consumption records of a household. The features then serve as the input to the  prediction model. Table \ref{priv:num_sim:feature:table} lists all 23 features we calculated from electricity consumption data, which is also used in \cite{beckel2014revealing}. We treat these features as $X$ and normalize them such that they range from 0 to 1. Data normalization is required in our experiment in that it gets rid of the scale inconsistency across the different features.%

%%%%%%%%%%%%%%%%%%%%%%%%%%%%%%%
% 
\begin{figure}[!ht]
      \centering % if in IEEE format, change the boxwidth=1.55 
      \framebox{\parbox[b]{1.51in} 
 {\includegraphics[scale=0.21]
 {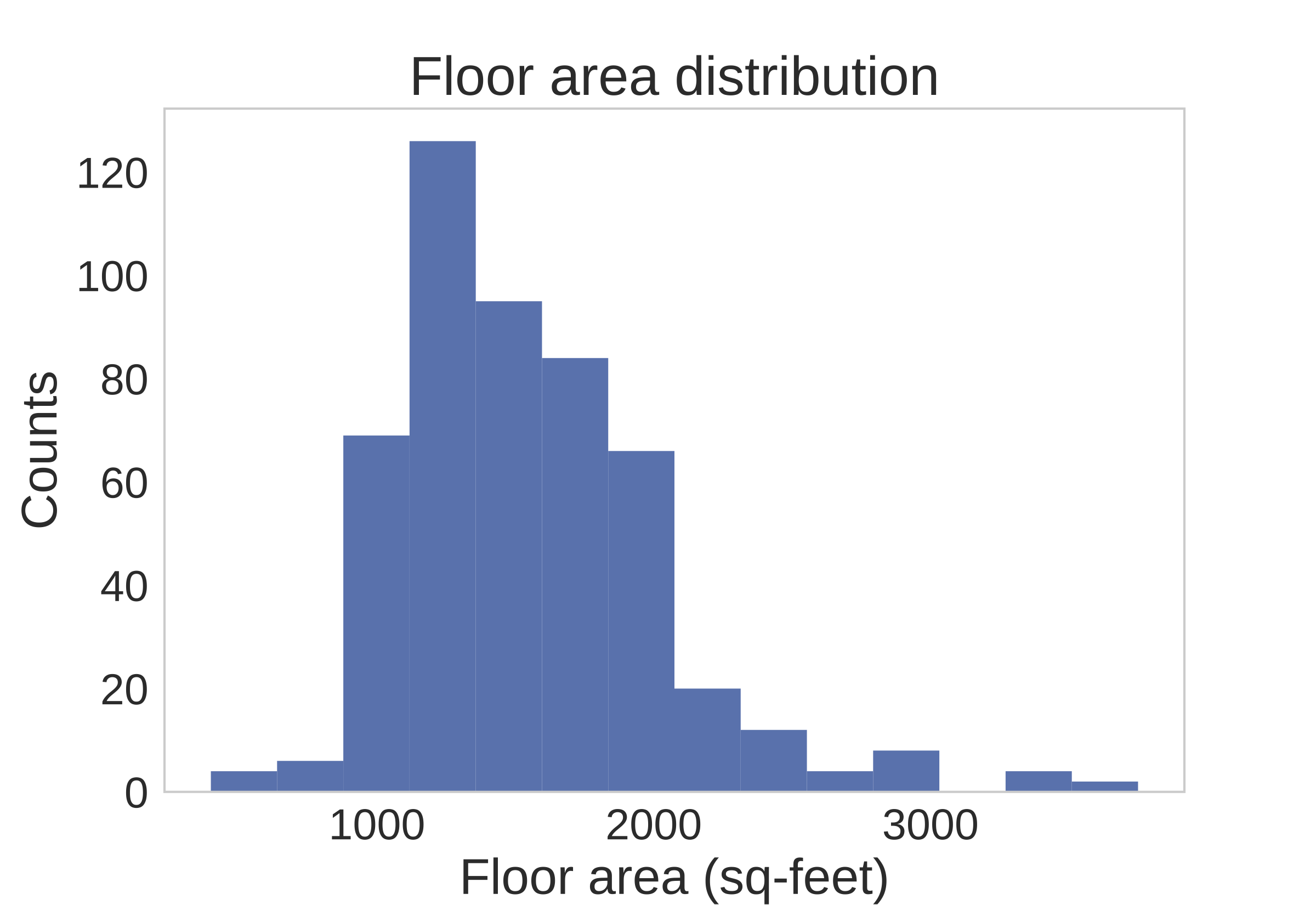} }{\includegraphics[scale=0.19]{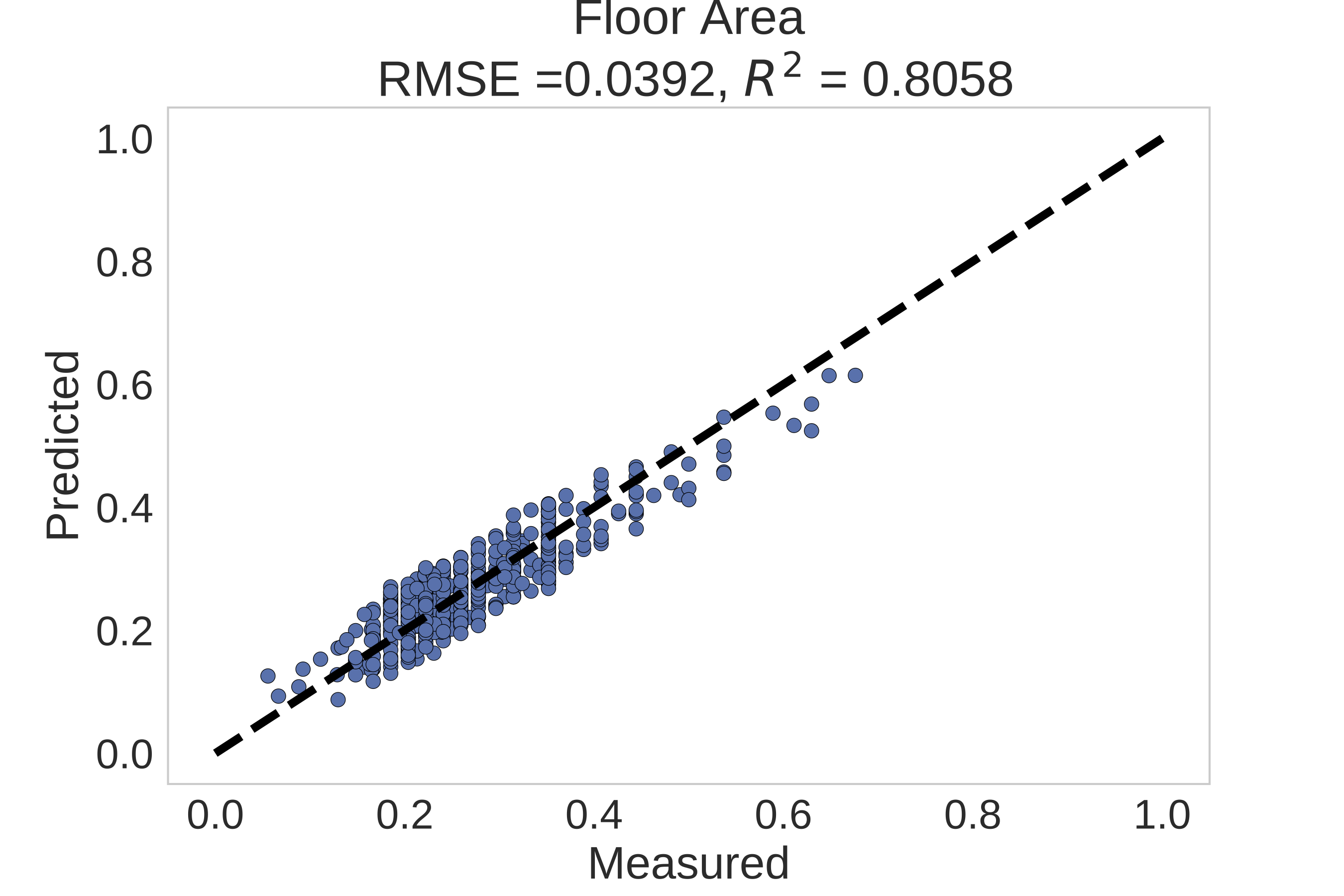}}  
 }
      \caption{Display the sampled data. \textbf{Left panel} shows the histogram of \emph{Floor area} of each house. \textbf{Right panel} shows the prediction capability of a linear model that fits on the \emph{Floor area}} %\emph{\textbf{Number of appliance}} per household; the right panel shows the histogram of
  \label{pes:priv:hist:n_fl_area}
\end{figure}
\begin{figure*}[!bpht]
\centering
\framebox{\parbox[b]{2.0in}
{\includegraphics[scale=0.26]{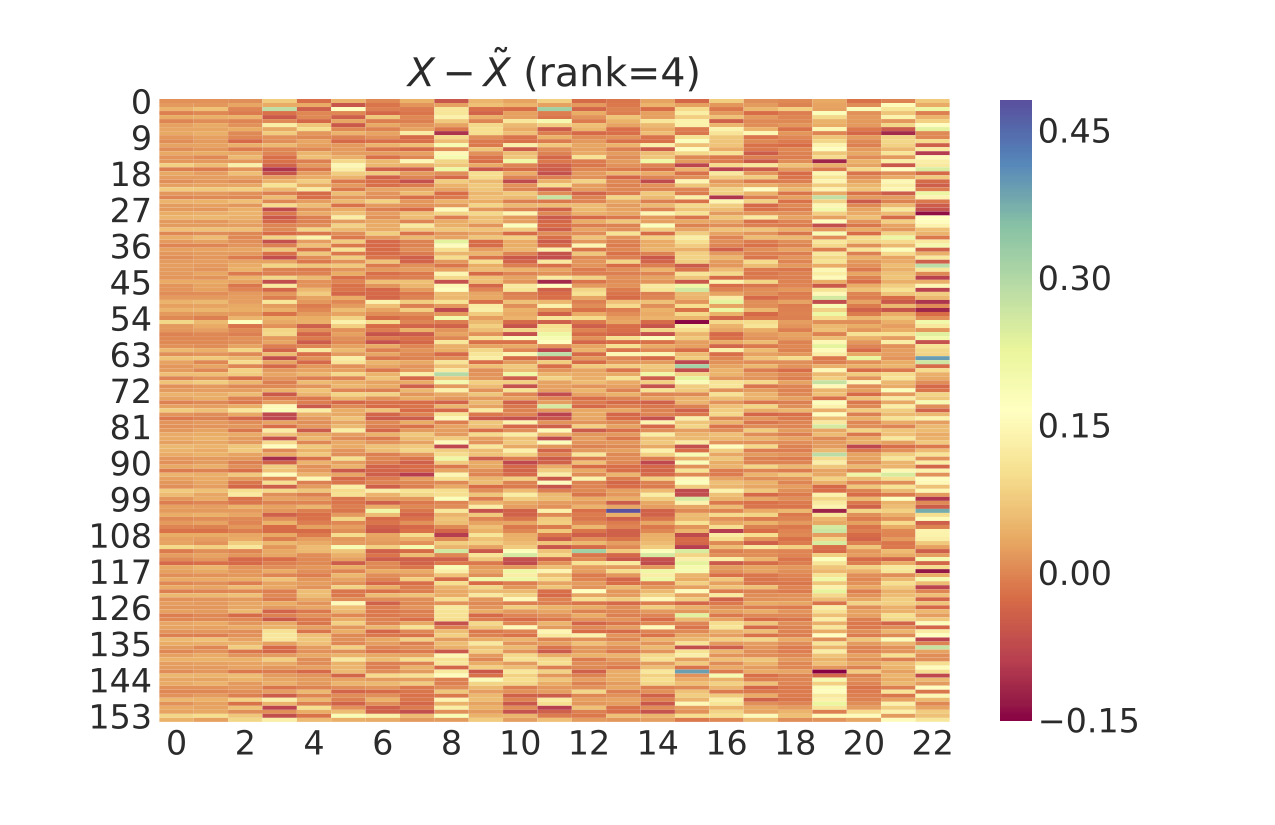}
}
{\includegraphics[scale=0.26]{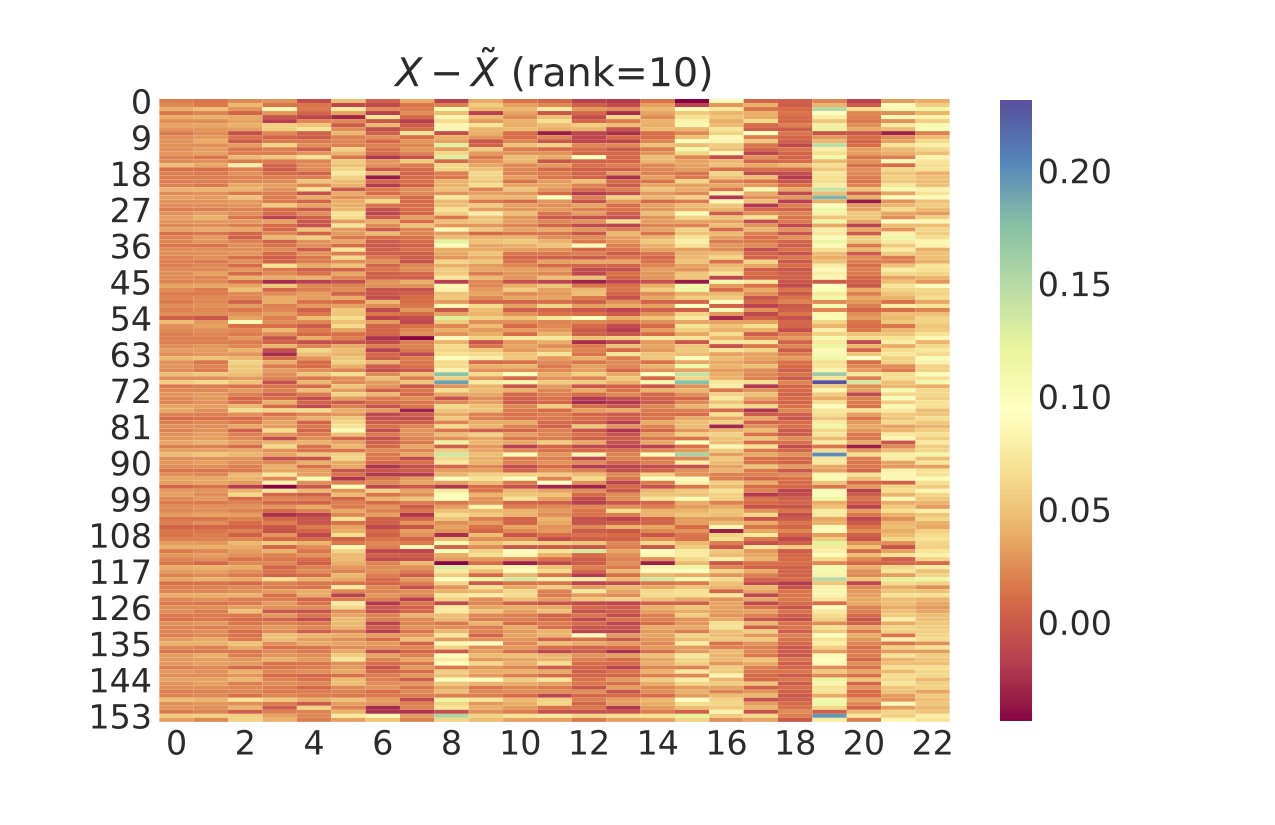}
}
{\includegraphics[scale=0.26]{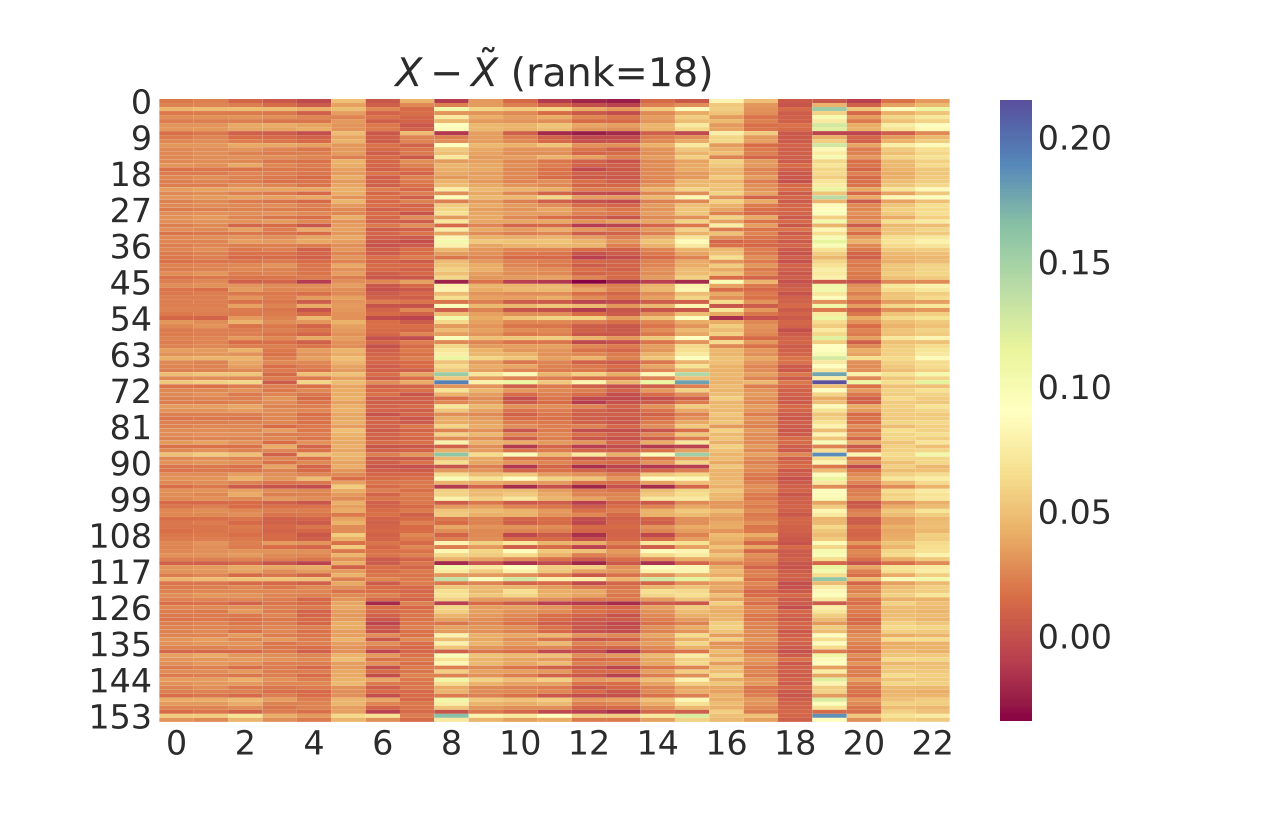}
}}

\caption{Difference between altered and original data when rank equals 4, 10, and 18}
\label{cdc:priv:linear:diff:batch:samples}
\end{figure*}

In the linear transformation model, given $Y$ is the private information, we run algorithm \ref{pes:priv:alg:linear:gen:X_tilde} to release $\tilde{X}$. This procedure involves solving semidefinite programming, which could be slow when the dimension of input samples is large. So we partition the samples into several groups with a reasonable number of households in each group (e.g. 30 to 40 as long as the number of households is larger than the number of features). After running experiments on several rank conditions of data matrices, we found that lower rank indicates better privacy (higher prediction error), given in Table~\ref{priv:table:linear:svd:statistics}. With a low rank condition that the data holder maintains, the attacker can barely (see Table~\ref{priv:table:linear:svd:statistics}) predict the private label $Y$. We also partitioned the data into 80\% for training and 20\% for testing. Table \ref{priv:table:linear:svd:statistics} shows the corresponding results with different ranks of the compression matrix for the testing set. A batch of released data differs from the original when rank is $4, 10$ and $18$. The difference is shown in Figure~\ref{cdc:priv:linear:diff:batch:samples}.  

%
%
%
% %---add on the sample visualization---%
% \begin{figure}[!ht]
% \vskip -0.12in
% % \begin{center}
% \framebox{\parbox[b]{1.49in}
% {\includegraphics[scale=0.20]{fig_linear/original_data_samples_heatmap.png}}
% \hspace{-0.01in}
% {\includegraphics[scale=0.20]{fig_linear/altered_data_rank_4_samples_heatmap.png}
% }}
% \framebox{\parbox[b]{1.485in}
% {\includegraphics[scale=0.204]{fig_linear/altered_data_rank_10_samples_heatmap.png}}\hspace{-0.01in}
% {\includegraphics[scale=0.204]{fig_linear/altered_data_rank_18_samples_heatmap.png}
% }}
% % \end{center}
% \vspace{-0.10in}
% \caption{The data holder applies a linear transformation on the data, i.e. compresses it down to low rank matrices and recovers it back to the original dimension. Each figure consists of 155 sampled households with 23 features which has been normalized originally. We fix the distortion tolerance $\gamma$ and tune the nuclear norm coefficient $\beta$ to get the result of various rank scenarios. }
% \label{priv:fig:linear:svd:multi_rank_cases:heatmap}
% \end{figure}
% % ----------------------------

%%%%%%%%%%%%%%%%%
% Linear result %
%%%%%%%%%%%%%%%%%
\begin{table}[!ht]
\vspace{-0.11in}
\centering
\caption{Metrics of Linear transformed data}
\label{priv:table:linear:svd:statistics}
\begin{tabular}{cccc}
\hline
Rank & RMSE & $R^2$ & distortion  \\ 
\hline
4  & 6.7e+04 & 1.61e-08 & 0.616 \\ %0.02679*23
10 & 7.1e+01 & 2.28e-03 &  0.081\\ %0.00345*23 \\
18 & 8.7e-01 & 1.12e-02 & 0.079 \\ %0.00345 * 23
23 & 3.9e-02 & 8.01e-01 & 0 \\
\hline
\end{tabular}
\end{table}
\vspace{-0.19in}
%
%%%%%%%%%
%

%% file: priv_pgf_autoencoder.tex
% \begin{figure}[!hpbt]
% \centering	
% \begin{tikzpicture}[node distance=1cm, auto, thick,scale=0.8, every node/.style={transform shape}]
% 	\node[punkt] (origdata) {Original\\ Data};
% 	\node[punkt, right=of origdata] (compressdecompress) {Encoder-Decoder}
% 	     edge[pil, <-] (origdata.east) ;  
% 	\node[punkt, right=of compressdecompress] (altData) {Altered Data}
% 	     edge[pil, <-] (compressdecompress.east);
%     \node[punkt, below=of compressdecompress](attacker){Attacker} 
%     edge[pil, <-](altData.south)
%     edge[pil, ->](compressdecompress.south);
% \end{tikzpicture} 
% \caption{Data compression schematic. The encoder and decoder can be linear or nonlinear}
% \end{figure}

\begin{figure}[!hbpt]
\begin{center}
\begin{tikzpicture}[auto, node distance=1.2cm,>=latex', thick, scale=0.9, every node/.style={transform shape}]
\node [input, name=input] {};
\node [block, right =of input](LinEncoderDecoder){ Compress \& Reconstruct};
% [,pin = {[pinstyle]above:$\varepsilon$(optional)}]
\node [blockAtt, right = of LinEncoderDecoder] (Attacker) {Attacker $\hat{Y} = h(\tilde{X}, Y)$ };
\node [output, right = of Attacker] (outputLoss) {}; 
\draw [draw, ->](input) -- node {$X$} (LinEncoderDecoder);
\draw [draw, ->] (LinEncoderDecoder) -- node {$\tilde{X}$} (Attacker);
\draw [draw, ->] (Attacker) -- node[annot, name=attLoss] {$\ell(\hat{Y}, Y)$} (outputLoss) {};
\path [draw, ->] (attLoss) -- +(0, -1.5) -| (LinEncoderDecoder); 
\end{tikzpicture}
\end{center}
\caption{Data compression schematic. 
%$\varepsilon$ is the random noise that convolve with compressed data $X$. 
$\tilde{X}$, which has the same dimension as $X$, is the reconstructed data that will be released to the public. The attacker infers the private labels $Y$ by choosing good predictor $h$, getting the resulting $\hat{Y}$, and minimizing the inference loss denoted as $\ell(\hat{Y}, Y)$.}
\label{cdc:fig:compress:diag}
\end{figure}

%% file: cdc_priv_data_game_nonLinear.tex
\subsection{Nonlinear compression with categorical variable}
Another common type of data has publishable features $X$ are high-dimensional continuous and the private labels $Y$ are discrete, for instance, images with some discrete labels (e.g. gender). Generally speaking, a sample $i$ has $y_i \in \mathcal{Y}=\{-1, +1\}$ and $x_i \in \mathbb{R}^p$ where $x_i^T$ is the $i$th row of the data matrix $X$. The data holder designs a nonlinear compression mechanism to reduce the classification accuracy of $y_i$ given $\tilde{x}_i$, where $\tilde{x}_i=g(x_i, y_i)$. We assume the attacker can use an advanced model, e.g. neural networks, to estimate the private labels. We further specify that $h$ and $g$ are functions parametrized by $\theta_h$ and $\theta_g$. The attacker minimizes the estimation loss, that is, $\min_{\theta_h}\ell(h_{\theta_h}(g_{\theta_g}(X, Y)), Y)$. The data holder designs a compressive function $g$ to maximize the attacker's loss as well as maintain a certain distortion $\gamma$ as aforementioned in equations (\ref{cdc:prob:formulation:general1}),(\ref{cdc:prob:formulation:general2}). This min-max game is difficult to find its equilibrium point in the context of neural networks with constraints, because the objective functions are non-convex with respect to parameters. Therefore, we use a heuristic way to cast the constrained optimization into a unconstrained optimization with regularization as follows: 
% \par{\small
%  \setlength{\abovedisplayskip}{1pt}
%  \setlength{\belowdisplayskip}{\abovedisplayskip}
%   \setlength{\abovedisplayshortskip}{0pt}
%   \setlength{\belowdisplayshortskip}{1pt}
\begin{align}
\max_{\theta_g}\Big\{\min_{\theta_h}\frac{1}{n}\sum_{i=1}^n\ell\Big(h_{\theta_h}\big(g_{\theta_g}(x_i, y_i)\big), y_i\Big)  \\
- \beta\big( (\frac{1}{n}\sum_{i=1}^n\| g_{\theta_{g}}(x_i, y_i) - x_i\|^2) - \gamma \big)^2  \\
+ \rho \min\{0, \gamma - (\frac{1}{n}\sum_{i=1}^n\| g_{\theta_{g}}(x_i, y_i) - x_i\|^2) \} \Big\}, \label{cdc:prob:nonlinear:minmax:augment:term2}
\end{align}
% }
where $\beta$ and $\rho$ are the hyper parameters controlling the iterates satisfied by the constraints. The distortion is characterized by the averaged Euclidean norm of the difference in samples. We propose a simple min-max alternative algorithm (Algorithm \ref{cdc:priv:alg:nn:gen:X_tilde}) to obtain the parameter $\theta_g$ for the function $g$ and yield the corresponding $\tilde{X}$.
\begin{algorithm}[!ht]
\caption{Generating $\tilde{X}$ (Neural Net attacker)}
\label{cdc:priv:alg:nn:gen:X_tilde}
\begin{algorithmic}[1]
\STATE \textit{Input:} dataset $\mathcal{D}$,  parameter $\gamma$, iteration number $T$
\STATE \textit{Output:} Optimal data publisher parameters $\theta_{g}$ 
% 		\Function {Alternate Minimax}{$\mathcal{D}, \gamma, T$}
		\STATE Initialize $\theta^t_{g}$ and $\theta^t_{h}$ when $t=0$
		\FOR{$t=0,...,T$} 
		\STATE take minibatch of $n$ samples $\{x_{(1)}, \hdots,x_{(n)}\}$ drawn randomly from $\mathcal{D}$
		\STATE Generate $\tilde{x}_{(i)}=g_{\theta_g}(x_{(i)},y_{(i)})$ for $i=1,\hdots,n$ 
		\STATE Compute the parameter $\theta^{t+1}_h$ for the adversary	{$\quad \theta^{t+1}_h=\arg\min_{\theta_{h}}\frac{1}{n}\sum\limits_{i=1}^{n}\ell(h_{\theta_h}(\tilde{x}_{(i)}),y_{(i)})$}
		\STATE Compute the descent direction $\nabla_{\theta_{g}} \mathcal{L}(\theta_{g}, \theta^{t+1}_h)$, where
		{$$
		\nonumber
\mathcal{L}(\theta_{g},\theta^{t+1}_h)= -\frac{1}{n}\sum\limits_{i=1}^{n}\ell(h_{\theta^{t+1}_h}(g_{\theta_g}(x_{(i)},y_{(i)})),y_{(i)}) $$}{$$ 
+ \beta\big( (\frac{1}{n}\sum_{i=1}^n\|g_{\theta_g}(x_{(i)}, y_{(i)}) - x_{(i)}\|_2^2) - \gamma \big)^2 $$} 
{$$ + \rho \max\{0,  (\frac{1}{n}\sum_{i=1}^n\| g_{\theta_{g}}(x_i, y_i) - x_i\|^2) - \gamma \}
		$$}
		\STATE Perform backtracking line search along $ \nabla_{\theta_{g}} \mathcal{L}(\theta_{g}, \theta^{t+1}_h)$ and update
		{$\theta^{t+1}_{g}= \theta^{t}_{g}-\alpha_t \nabla_{\theta_{g}} \mathcal{L}(\theta_{g}, \theta^{t+1}_h), \quad\alpha_t>0$}
		\STATE Exit if solution converged
		\ENDFOR
		\STATE \textbf{return} $\theta^{t+1}_g$; 
% 		\EndProcedure
$\tilde{X} = g_{\theta^{t+1}_g}(X, Y)$
\end{algorithmic}
\end{algorithm}
Similar to the idea of the Augmented Lagrangian method\cite{wu2010augmented}, the scale of $\beta$ and $\rho$ are gradually increasing as the iteration step increases. The term (\ref{cdc:prob:nonlinear:minmax:augment:term2}) is added to ensure the solution strictly satisfies the constraint mentioned in expression (\ref{cdc:prob:formulation:general2}). Other alternative approaches are also proposed in \cite{goodfellow2014generative,hamm2016minimax,huang2017context}. Distinguished from those works, we construct a convex approximation with distortion constraints that is applied in privacy games.      
%
%
%
%
% the loss function is logistic loss $\ell_{\theta}(h(\tilde{x}), y) = \sum_{i=1}^N \log(1+\exp(-y_i\theta^T\tilde{x_i})) $. 
% The compression operation is performed such that $\tilde{x}_i = AA^{\dagger}{x}_i$ where $A \in R^{p\times k}$ and $A^{\dagger} \in R^{k\times p}$ given the result in equation \ref{linear:sudo:inverseA}. 
%
%
%%%%%%%%%%%%%%%%%%%%%%%%%%
% illustrating results in the case when y label are binary 
%%%%%%%%%%%%%%%%%%%%%%%%%%
\subsection{Case study: Images of people}
To perform our experiment of the nonlinear compressive model with a categorical response variable, we use the Groups of People dataset\cite{gallagher_cvpr_09_groups}. The dataset contains 4550 images from Flicker of human faces with labeled attributes such as age and gender. These images are $61 \times 49$ in grayscale pixels ranging from 0 to 255, with 3500 training and 1050 testing samples respectively. In this experiment, the images are $X$ and the label of gender, which is evenly spread in both the training and testing sets, is $Y$. We label female or male as 1 or -1. Sampled raw images are shown in Appendix \ref{cdc:appx:original:sampled:imgs}. %

For the data holder to perform nonlinear compression, we implement a three-layer neural network, which shares the similar concept of the autoencoder\cite{hinton2006reducing}. The first two layers serve as an encoder. The initial layer has 2989 units that takes original vectorized images [$2989=61 \times 49$], followed by a ReLU activation and batch normalization. We vary the second layer units from $2048, 512$, and $128$ for several cases, which are denoted as compression-rank. We define the corresponding compression-rank rate 0.685, 0.171, and 0.043 to be \emph{high}, \emph{medium} and \emph{low} respectively\footnote{the compression-rank rate is obtained by number of bottleneck units divided by input units. e.g. $\frac{2048}{2989} = 0.685$}. The last layer, connected with ReLU activation, has the same dimension as the vectorized image input that performs the role of a decoder.  
The attacker is represented by a 3-layer neural network, comprised of an initial 2989 units layer, followed by 2048 units layer, and lastly a two units layer as softmax output. We apply leaky ReLU activation and batch normalization between each layer. %
  
\begin{figure}[!ht]
\begin{center}
\centerline{\hspace{0.0in}
\includegraphics[width=0.23\columnwidth]{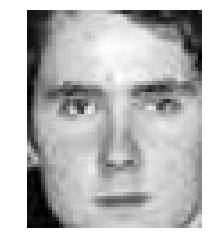}
\hspace{-0.2in}
\includegraphics[width=0.23\columnwidth]{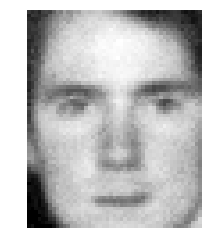}
\hspace{-0.2in}
\includegraphics[width=0.23\columnwidth]{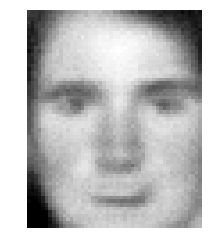}
\hspace{-0.2in}
\includegraphics[width=0.23\columnwidth]{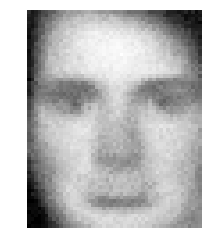}
}
\centering{\includegraphics[width=0.3\columnwidth]{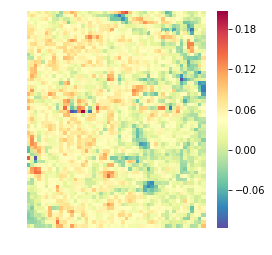}
\hspace{-0.2in}
\includegraphics[width=0.3\columnwidth]{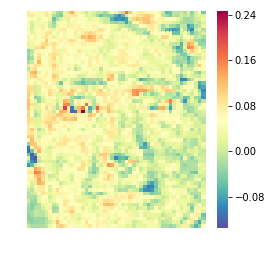}
\hspace{-0.2in}
\includegraphics[width=0.3\columnwidth]{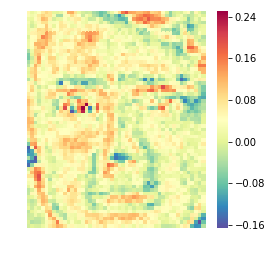}
}
\caption{A sampled face. \textbf{Top row}: From left to right are an original image, and then decoded images with \emph{high, medium} and \emph{low} compression-rank rate. \textbf{Bottom row}: From left panel to right panels are the result of the pixel difference between decoded and raw image projected to 0-255 with \emph{high}, \emph{medium} and \emph{low} compression-rank rates respectively}
\label{cdc:fig:img:wo:gan:rank}
\end{center}
\end{figure}

Before considering adversarial compression, we first classify reconstructed images with different compression-rank rates without having a min-max game. This operation serves two purposes: a) investigating the accuracy of gender classification; b) fetching the minimum distortion threshold $\gamma$ in the context of mean squared error loss (i.e. min $\frac{1}{n}\sum_{i=1}^{n} \|\tilde{x}_i - x_i\|^2$ yields the smallest $\gamma$). The following results are evaluated based on the testing set. Figure~\ref{cdc:fig:img:wo:gan:rank} displays a sampled image associated with different scenarios. A lower compression rank rate yields a worse image quality. Table \ref{cdc:table:face:compress:wo:gan:clf:res} shows that compressing images with the high and medium ranks doesn't reduce the gender classification accuracy too much, yielding a relatively low image quality loss. In the example of high compression-rank, the average distortion per pixel is $0.0166 * 10^{2} \approx 1.6 \% $ which is not too large.    
\begin{table}[!ht]
\vspace{-0.15in}
\caption{Classification results of gender with the raw data under different compression-rank cases}
\label{cdc:table:face:compress:wo:gan:clf:res}
\centering
\begin{tabular}{cccc}
\hline
compression-rank &  accuracy(gender) & distortion/pixel & distortion \\
\hline
raw (2989) &   0.692   & 0 & 0\\
high (2048) &  0.685   & 0.0166 & 0.195 \\
medium (512) &  0.664   & 0.0259 & 0.304 \\ 
low (128) &  0.627 & 0.0312 & 0.365\\
\hline 
\end{tabular}
\end{table}
\vspace{-0.05in}
\par
\begin{figure}[!ht]
\centerline{
\hspace{-0.12in}
\includegraphics[width=0.17\columnwidth]{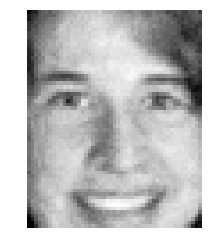}
\hspace{-0.18in}
\includegraphics[width=0.17\columnwidth]{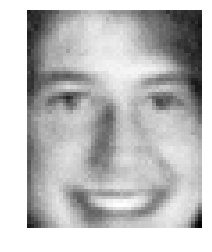}
\hspace{-0.18in}
\includegraphics[width=0.17\columnwidth]{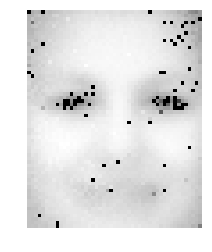}
\hspace{-0.18in}
\includegraphics[width=0.22\columnwidth]{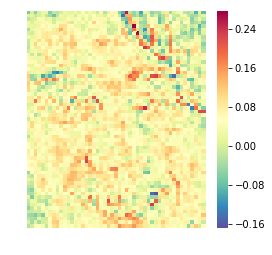}
\hspace{-0.2in}
\includegraphics[width=0.22\columnwidth]{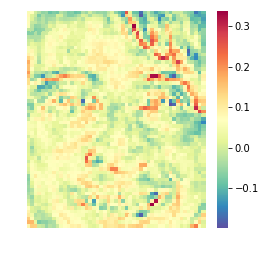}
\hspace{-0.2in}
\includegraphics[width=0.22\columnwidth]{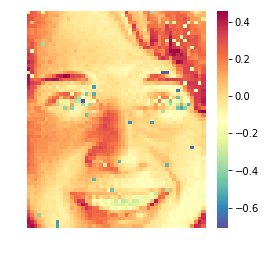}
}
\centerline{
\hspace{-0.11in}
\includegraphics[width=0.17\columnwidth]{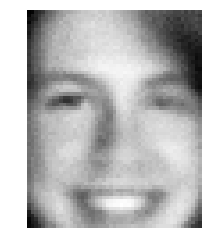}
\hspace{-0.18in}
\includegraphics[width=0.17\columnwidth]{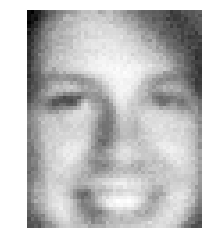}
\hspace{-0.18in}
\includegraphics[width=0.17\columnwidth]{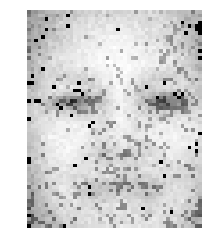}
% }
% \centerline{
% ~
\hspace{-0.2in}
\includegraphics[width=0.22\columnwidth]{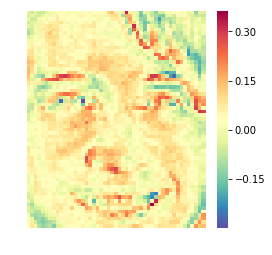}
\hspace{-0.2in}
\includegraphics[width=0.22\columnwidth]{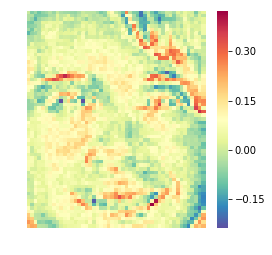}
\hspace{-0.2in}
\includegraphics[width=0.22\columnwidth]{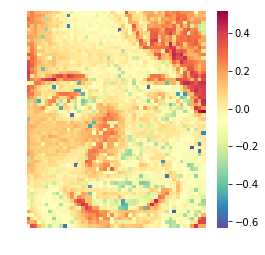}
}
\caption{A sampled image with different compression rank and distortion tolerance $\gamma$. \textbf{Top left}: From left to right are visual output when $\gamma$ equals 0.3, 2, 4 in the high compression-rank case. \textbf{Top right}: From left to right, these images show the difference between output images and raw images for corresponding $\gamma$ value in the high compression-rank case. \textbf{Bottom left}: From left to right are visual output when $\gamma$ equals 1, 2, 4 in the low compression-rank case. \textbf{Bottom right}: From left to right are difference between output images and raw images for corresponding $\gamma$ value in the low compression-rank case.}  
\label{cdc:fig:altered:face:high:low:imgs}
\end{figure}

{
\setlength\intextsep{0mm}
Utilizing the previous result as a reference, we pick several proper values of $\gamma$ to further understand the adversarial privacy compression. In the high compression-rank case, we test three scenarios where $\gamma$ is $0.3, 2$ and $4$ respectively. We discover that the encoder-decoder tends to alternate pixels near eyes, mouths, and rims of hair. A similar patten can also be observed when we test the low compression-rank case where $\gamma$ is $1,2$ and $4$. We also notice that the low compression-rank scenario has a more scattered dotted patten of black/gray pixels at the large tolerance level, whereas the high compression-rank case has more concentrated black pixels, as shown in Figure~\ref{cdc:fig:altered:face:high:low:imgs}. We believe the reason is that the data holder always adjusts the pixels that are highly correlated with gender. Since the high compression-rank encoder-decoder preserves more information than the low compression-rank one, it's much easier for the data holder to alter the target pixel features within limited total distortion. The privatized images generated through min-max training indeed yield lower prediction accuracy of gender than the original encoded-decoded images. Table \ref{cdc:table:face:compress:gan:clf:res} depicts the gender classification results indicating that it is harder to predict gender with increased distortion. The table also reveals that higher compression rank performs better in terms of decreasing the accuracy if the distortion is sufficiently large.  }             % 
\begin{threeparttable}[!hbpt]
\vspace{-0.05in}
\caption{Classification accuracy of gender with the data released under adversarial privacy}
\label{cdc:table:face:compress:gan:clf:res}
\centering
\begin{tabular}{ccccc}
\hline
compression-rank & $\gamma=0.3$ & $\gamma=1$ & $\gamma=2$ & $\gamma=4$ \\
\hline
high (2048) & 0.628 & 0.600 & 0.573 & 0.486 \\ 
medium (512)\tnote{1} & & 0.607 & 0.594 & 0.512 \\
low (128)  & & 0.602 & 0.585 & 0.521  \\
\hline
\end{tabular}
\begin{tablenotes}
    \item[1] \small{$\gamma=0.3$ is unattainable, since the compression rank is small enough so that the minimum reconstruction loss (Mean Squared Error) already reachs to the 0.3.}
\end{tablenotes}
\end{threeparttable}

%% file: priv_data_game_thm_guarantee.tex
\section{Privacy Guarantee}
\label{cdc:quantify:priv:garentee:s4}
%add content on computing mutual information
%%%%%%%%%%%% add transition on why use MI to measure the correlation %%%%%%%%%%%%%%%
Our previous experiments show that a (local) equilibrium can be achieved through this min-max game approach. While we cannot preclude that there may be some other equilibria in the context of a neural network. Thus, a quantifiable metric is needed to give privacy guarantee between sensitive data and altered data.      

In this section we introduce the empirical  mutual-information concept to quantify the privatization quality of this min-max game approach, i.e. measuring the correlation between the sensitive response data and the released feature data pre- and post-privatization. Mutual information
(MI) \cite{shannon2001mathematical} is a well established tool that has been widely adopted to quantify the correlation between the two streams of data by a non-negative scalar \cite{cover2012elements}. 
%The mutual information between random variables $X$ and $Y$ is known as $I(X; Y) = H(X) - H(X|Y)$, where $H(X)$ is the entropy of random variable $X$ and $H(X|Y)$ is the conditional entropy of random variable $X|Y$. 
From the data driven perspective, we have the empirical MI $\hat{I}(X; Y) = \hat{H}(X) - \hat{H}(X|Y)$, where $\hat{H}$ characterizes the empirical entropy. This empirical entropy can be calculated using the classical nearest $k$-th neighbor method \cite{kraskov2004estimating}.
% \begin{align}
% \hat{H}(X) = \psi(N) - \psi(k) + \log(c_d) + \frac{d}{N}\sum_{i=1}^N\log r_i \label{priv:garentee:entropy:est:knn}
% \end{align}
% where $r_i$ is the distance of the $i$-th sample $x_i$ to its $k$-th nearest neighbor, $\psi$ is the digamma function, $c_d = \frac{\pi^{d/2}}{\Gamma (1+d/2)}$ in Euclidean norm or $c_d=2^d$ in maximum norm, and $N$ is the number of samples.

% \subsection{Continuous response label}
\textbf{Continuous response label}: Given that $Y$ is continuous, the mutual information can be expressed as 
% \begin{align}
$\hat{I}(X;Y) = \hat{H}(X) + \hat{H}(Y) - \hat{H}(X,Y)$,
% \end{align}
where $\hat{H}(X), \hat{H}(Y)$ can be obtained directly from method in \cite{kraskov2004estimating},
% equation (\ref{priv:garentee:entropy:est:knn})
given samples $x_i, y_i$. The joint empirical entropy is calculated by concatenating each $x_i$ and $y_i$ together as one sample and using the nearest neighbor entropy estimation again.

%%%%%%%%%%%%%%%%%%
% \subsection{Categorical response label}
\textbf{Categorical response label}: 
For the discrete response $Y \in \{-1, +1\}$, we have the mutual information
%\begin{align*}
$\hat{I}(X;Y)= \hat{H}(X)-\hat{H}(X|Y) = \hat{H}(X)- \big(p(Y=-1)\hat{H}(X|Y=-1) + p(Y=1)\hat{H}(X|Y=1) \big)$,
% \end{align*}
where $p(Y=\pm 1)$ can be approximated by the sample frequency in the dataset, and $\hat{H}(X|Y=\pm 1)$ can be obtained by the aforementioned $k$-th nearest neighbor method with partitioned samples according to the value of $Y$. 

% \subsection{Quantifying MI}
For the experiment of the continuous response variable, we first calculate the empirical MI between the power consumption statistics and floor areas. The original MI between power usage statistics data and floor area is $\hat{I}(X;Y) = 2.150$. The resulting MI between altered power usage data and floor areas, which is denoted by $\hat{I}(\tilde{X};Y)$, are 0.995, 0.494, and 0.216 when the rank of compression matrices are $18$, $10$, and $4$. For the categorical response variable experiment, the empirical MI between the images data and gender data is obtained as follows. The original MI between raw images $X$ and gender label $Y$ is $\hat{I}(X;Y) =0.249$. When we pick high compression-rank with $\gamma=0.3, 1, 2$ and $4$, the $\hat{I}(\tilde{X}; Y)$ are $0.217, 0.170, 0.105$, and $0.012$. The medium compression-rank yields $\hat{I}(\tilde{X}; Y)$ to be $0.179, 0.112$, and $0.014$ for $\gamma = 1, 2,$ and $4$ respectively. In the low compression rank case, $\hat{I}(\tilde{X};Y)$ are $0.174, 0.101,$ and $0.017$ with the aforementioned $\gamma$. We notice the empirical MI indeed decreases as the distortion increases. Due to the challenge of high dimensional data, we apply the principal component analysis to project the $\tilde{X}$ down to 16 dimensions and find the approximate $\hat{I}(\tilde{X};Y)$. This is an alternative attempt to demonstrate the effectiveness of using our framework. The changes of mutual information value show the privacy guarantee between the released data and sensitive labels under various distortion conditions. Yet we believe a more advanced architecture of the neural network can be applied to extract the embeddings of semantic features, resulting a better estimates of empirical mutual information. We will explore this potential direction in our future research. 

%
%Part of the reason is the intrinsic approximation error when estimating the empirical entropy. Another reason is that the low rank approximation of the original data reduces the intrinsic variability of the data itself, which results in having rigid and consistent values that are realized by $\tilde{X}$. This could ultimately increase the empirical mutual information between $\tilde{X}$ and $Y$. 
% 
% believe incorporating mutual information in the min-max training procedure may help understanding how mutual information relates to compression rate, and it could be a potential direction for future research.

%
% \begin{table}[!bpht]
% \caption{$\hat{I(X; Y)}$}
% \label{}
% \begin{tabular}{cccccc}
% \hline
% compression-rank &  $\gamma=0.3$ & $\gamma=1$ & $\gamma=2$ & $\gamma=4$\\
% \hline 
% % raw (2989) & 166.975 &  &  &  & \\
% high (2048) &     & 132.037  & 121.928  & 110.33 &  \\
% medium (512) &  & &  &  & \\
% low (128)  &   & &  &  & \\
% \hline 
% \end{tabular}
% \end{table}

%\\~\\ 

%% file: priv_data_game_conclusion_discuss.tex
\section{Conclusion}
\label{cdc:conclusion:s5}
% rewrite conclusion 
Recent breakthroughs in artificial intelligence require a huge amount of data to support the learning quality of various models. Yet the risk of data privacy is often overlooked in the current data sharing processes. The recent news of data leakage by Facebook shows that privacy risk could significantly impact some issues in politics. Therefore, securely designing a good data privatization mechanism is important in the context of utilizing machine learning models. By thorough evaluations, our new min-max adversarial compressive privacy framework provide an effective and robust approach to protect private information. We leverage the data-driven approach without posing assumptions on data distribution. It's crucial during practical implementation, since the real data is often more complicated than a simple characterization of a parametric probability distribution. Along this line of research, many interesting extensions can be built on our framework to create a robust privacy protector for data holders. 

%Our framework can  various attackers in realistic scenarios, and demonstrate that our approach protects sensitive information robustly from any data holder's perspective.             

%%%%%%%%%%%%%%%%%%%%%%
% Privacy concerns have drawn a large amount of attention in current modeling and algorithm design. This paper presents a thorough evaluation of the min-max adversarial privacy framework which is proposed at the beginning of the paper, showing that a data holder indeed can use such a framework to release data with a certain trade-off between data privacy and utility. We extend the framework to compression scenario, namely compressive adversarial privacy framework. It leverages the data-driven approach without posing assumptions on data distribution. In compressive adversarial privacy framework, we show that some special case can have equilibrium for both attacker and data holder. We derived various algorithms and conduct realistic experiments on different scenarios, considering both linear and nonlinear compression under simple and sophisticated attacks. We demonstrate that this compressive context-aware adversarial privacy framework can give protections on sensitive information. %    

%% file: priv_data_g_appendix.tex
% \vspace*{-0.15in}
\section{Appendix}
%%%%%%%%%%%%%%%%%%%%%%%%%
% \vspace*{-0.14in}
%
\subsection{Casting linear problem}
% \vspace*{-0.11in}
\label{priv:icml2018:cast:linear:implicit:solve}
Consider the following problem 
\begin{align}
\min_{B} \|XAB - X\|_F^2, 
\end{align}
where matrix $X$ is $n\times p$, matrix $A$ is $p \times k$, and matrix $B$ is $k \times p$. We give a brief minimizer derivation as follows: 
\begin{align*}
\|XAB - X\|_{F}^2 = \mathbf{Tr}\Big( (XAB - X)(XAB - X)^T \Big) \\ 
= \mathbf{Tr}\Big( XABB^TA^TX^T - 2XABX^T + XX^T\Big) .
\end{align*}
The derivative of the first term with respect to $B$ is  
\begin{align*}
\frac{\partial}{\partial B}\mathbf{Tr} (XABB^TA^TX^T )   = \frac{\partial}{\partial B}\mathbf{Tr} (BB^TA^TX^TXA) \\ = (A^TX^TXA) B + (A^TX^TXA)^T B = 2 (A^TX^TXA) B .
\end{align*}
The derivative of the second term with respect to $B$ is 
\begin{align*}
\frac{\partial}{\partial B}\mathbf{Tr}(2XABX^T) = 2\frac{\partial}{\partial B}\mathbf{Tr} (B X^TXA) = 2 (X^TXA)^T 
\end{align*}
Thus, we set the derivative equals zero and obtain the minimizer $B$ as follows: 
\begin{align}
\frac{\partial}{\partial B } \mathbf{Tr}\Big( XABB^TA^TX^T - 2XABX^T + XX^T\Big) \\
=  2 (A^TX^TXA) B - 2 (X^TXA)^T  \triangleq 0  \\
\implies (A^TX^TXA) B = A^TX^TX \\
\implies B = (A^TX^TXA)^{-1}A^TX^TX .
\end{align}
Now we design the $\tilde{X}$ such that 
\begin{align}
\tilde{X} = XAB = XA\underbrace{(A^TX^TXA)^{-1}A^TX^TX}_{B} \\
= X\underbrace{A(A^TX^TXA)^{-1}A^T}_{M}X^T X .
\end{align}
Instead of explicitly designing a low rank matrix $A$ , we solve an alternative equivalent problem of determining a low rank matrix ${M}$ to compress the data. 
% \vspace*{-0.17in}
\subsection{Recovering Linear Operation} 
\label{priv:icml2018:appendix:recover:mat:B}
% \vspace*{-0.12in}
Claim: $B$ is pseudoinverse of $A$, i.e. $B = (A^TA)^{-1}A^T = A^{\dagger} $. 

% Unless the case that data matrix $X$ and projection matrix $A$ has same dimension, say $n\times n$.  
\proof We apply Singular Value Decomposition (aka SVD, which is similar to PCA) on data matrix $X = USV^T$. $U$ is $n\times k$, $S$ is $k \times k$, $V$ is $p \times k$. We have 
\begin{align*}
B & =(A^TX^TXA)^{-1}A^TX^TX \\
& = \big(A^T(USV^T)^T(USV^T)A \big)^{-1} A^T (USV^T)^T(USV^T)
 \\
 & = (A^TVSU^TUSV^TA)^{-1} A^T VSU^TU SV^T \\
 & = (A^TVS\underbrace{U^TU}_{I} SV^TA)^{-1} A^T VS\underbrace{U^TU}_{I} SV^T \\
& = (A^TVS^2V^TA)^{-1} A^T VS^2V^T \\
& = (\underbrace{S^2A^TV}_{\text{swapped }}V^TA)^{-1} \underbrace{ S^2 A^TV}_{\text{swapped}}V^T \\
& = (S^2 A^T\underbrace{VV^T}_{I}A)^{-1} S^2 A^T \underbrace{VV^T}_{I} \\
& = (S^2A^TA)^{-1} S^2 A^T = 
% (A^TA)^{-1} S^{-2}S^2 A^T = 
(A^TA)^{-1} A^T .
\end{align*}
%%%%%%%%%%%%%%%%%%%
\subsection{Proof of positive semidefinite property of a matrix} \label{pes:priv:app:linear:tran:M:psd}
Claim: $M = A(A^TX^TXA)^{-1}A^T$ is Positive Semidefinite. 
\proof show $A^TX^TXA$ is positive semidefinite. Since $A \in \mathbb{R}^{p\times k}, X \in \mathbb{R}^{n \times p}$, for any vector $v \in \mathbb{R}^k$, we have 
\begin{align*}
v^TA^TX^TXAv = (XAv)^T(XAv) = \|XAv\|_2^2 \geq 0 .
\end{align*}
Therefore, we can apply Signaler Value Decomposition on $(A^TX^TXA)$, we get $(A^TX^TXA)= VSV^T$, where $S = \diag(\sigma_1, \sigma_2, ..., \sigma_k)$. The resulting $(A^TX^TXA)^{-1}$ can be expressed as 
$(A^TX^TXA)^{-1} = V(S^{-1})V^T$ where \\ 
$S^{-1} = \diag(1/\sigma_1, 1/\sigma_2, ..., 1/\sigma_k)$. Because all $\sigma$ are positive, we denote $\delta_i^2 = \frac{1}{\sigma_i}$. Hence $\Delta = \diag(\delta_1, ..., \delta_k)$. And 
\begin{align}
M = AV\Delta \Delta^T V^TA^T.
\end{align}
For any $v \in \mathbb{R}^p$, we have 
\begin{align}
v^T M v = v^T(\Delta^T V^TA^T)^T \Delta^T V^TA^T v \\
= \|\Delta^T V^TA^T v \|_2^2 \geq 0  .
\end{align}
Thus $M$ is positive semidefinite. 
%%%%%%%%%%%%%%%%%%%
%%%%%%%%%%%%
\subsection{Convexity of a re-parameterized problem} \label{pes:priv:app:linear:tran:M:cvx_prob1}
% Claim: $\min_{\tilde{M}} \big\{ Y^TXMX^TY + \gamma \|XMX^TX - X\|_{F}^2 \big\} $ is a convex problem. \\
Claim: The following optimization is convex: 
\begin{align}
& \min_{M} \Tr\big(C_{xy}^TX {M}X^TC_{xy}\big) + \beta \| {M}\|_{*}  \\
\text{s.t.} & \quad \|X{M}X^TX - X\|_{F}^2  \leq \gamma .
\end{align} 
%  & = Y^T\tilde{M}Y + \gamma \mathbf{Tr}\big\{ (\tilde{M} X - X)(\tilde{M} X - X)^T \big\}%
\proof It is easy to see that the first term $\Tr\big(C_{xy}^TX {M}X^TC_{xy}\big)$ is convex, since $M$ is positive semidefinite, trace operator is linear with respect to $M$. The second term and third term are also convex. For any norm, given $0 \leq \alpha \leq 1$ and two matrices $A, B$, we have \vspace*{-0.11in}
\par{\setlength{\abovedisplayskip}{0pt} \small{
\setlength{\abovedisplayskip}{1pt}
 \setlength{\belowdisplayskip}{\abovedisplayskip}
  \setlength{\abovedisplayshortskip}{0pt}
  \setlength{\belowdisplayshortskip}{1pt}
\begin{align*}
\|\alpha A + (1-\alpha) B\| & \leq \|\alpha A\| + \|(1-\alpha) B\| = \alpha\|A\| + (1-\alpha) \|B\|  
\end{align*}
}}
Hence Frobenius norm and Nuclear norm are specific forms of norm that is convex with respect to ${M}$. The first term in the objective is just linear in ${M}$. Thus, the problem is convex. 
%%%%
%%%%%%%%%%%%%%%%%%%%%%%%%
% if cut it off for page limits 
%%%%%%%%%%%%%%%%%%%%%%%%% 
% \begin{comment}
\subsection{Deriving linear compression with noise} \label{icml:appendix:linear:compress:noised:data}
Consider the case $\varepsilon \sim \mathcal{N}(0, \Sigma)$. We have the min-max game as follows
\begin{align}
& \max_{A, B, \Sigma}\min_{\Theta} \frac{1}{n}\sum_{i=1}^n \BBE_{\varepsilon \sim P_{\varepsilon} }  \|\Theta^T(A^Tx_i + \varepsilon) - y_i \|_2^2  \\
& - \gamma \|XAB - X\|_{F}^2
\end{align}
For attacker, we have the following minimization problem 
\begin{align}
& \min_{\Theta} \frac{1}{n}\sum_{i=1}^n \BBE_{\varepsilon \sim P_{\varepsilon} }  \|\Theta^T(A^Tx_i + \varepsilon) - y_i \|_2^2 \\
& = \min_{\Theta} \frac{1}{n}\sum_{i=1}^n 
\Big( (A^Tx_i + \varepsilon_i)^T \Theta \Theta^T (A^Tx_i + \varepsilon_i) \\
 & - 2 (A^Tx_i + \varepsilon_i)^T \Theta y_i + y_iy_i^T \Big)
\end{align}
We first find the minimizer $\Theta$ for the attacker. By taking the derivative over $\Theta$, we have 
\begin{align}
 \frac{1}{n}\sum_{i=1}^n\Big( A^Tx_ix_i^TA + \varepsilon_i \varepsilon_i^T\Big) \Theta = A^T \frac{1}{n} \sum_{i=1}^n x_i y_i^T \\
 \Theta = (A^TX^TXA + \Sigma)^{-1} A^T C_{xy}
\end{align}
Also we also find the best recover matrix $B$ by considering the following relation $\tilde{X} = XA + \mathbb{\boldsymbol{\varepsilon}}$
\begin{align}
& E_{\varepsilon \sim P_{\varepsilon}}\|\tilde{X} - X\|_{F}^2 \\
& = \Tr\Big((XAB + \boldsymbol\varepsilon B - X)(XAB + \boldsymbol\varepsilon B - X)^T\Big) \\
& = \Tr (XABB^TA^TX^T + B\Sigma B^T - 2XABX^T) 
\end{align}
Taking the derivative over $B$ and set it equals 0, we have $B = (A^TX^TXA + \Sigma)^{-1}A^TX^TX$. Hence the data holder's maximization can be casted into 
\begin{align}
& \min_{A, \Sigma} \Tr\Big(C_{xy}^TA(A^TX^TXA + \Sigma)^{-1}A^TC_{xy}\Big) \\
& + \|XA(A^TX^TXA + \Sigma)^{-1}A^TX^TX - X\|_F^2
\end{align}
It is not difficult to discover that $A(A^TX^TXA + \Sigma)^{-1}A^T$ is also positive semidefinite. Thus the problem can be relaxed to convex optimization.  
% \end{comment}
%%%%%%%%%%%%%%%%%%%%
\subsection{Proof of Theorem \ref{cdc:svd:err:th1}}
\label{cdc:appx:proof:Th1}
% \begin{proof}
\proof Denote $X = \sum_{i=1}^p \lambda_i u_i v_i^T$, where $\lambda_i$ is singular value, $u_i, v_i$ are corresponding left and right  singular vectors. The best rank-$k$ approximation $\tilde{X}_k = \sum_{i=1}^k \lambda_i u_i v_i^T$ is achieved by SVD in Frobenius norm by Eckart-Young theorem\cite{golub1987generalization}. Then $\|X-\tilde{X}_k\|_F^2 = \Tr\big( (\sum_{i=k+1}^p\lambda_i u_i v_i^T) (\sum_{i=k+1}^p\lambda_i u_i v_i^T)^T \big) = \Tr(\sum_{i=k+1}^p\lambda_i^2 ) = \sum_{i=k+1}^p \lambda_i^2$
% \end{proof}
%%%%
%%%%
%%%%
% \vspace*{-0.11in}
\subsection{Features extracted from power data}
% \vspace*{-0.15in}
%%%%%%%%%%%%%%%%%%%%%%%%
%% feature list table %% 
Features for power consumptions are displayed in Table~\ref{priv:num_sim:feature:table}.
\begin{table}[!hbpt]
\centering
\caption{Features extracted out of power consumptions}
\label{priv:num_sim:feature:table}
\begin{tabular}{|l|c|}
\hline 
Index & Description  \\ 
\hline 
1 & Week total mean \\
2 & Weekday total mean \\
3 & Weekend total mean \\
4 & Day (6am - 10pm) total mean \\
5 & Evening (6pm - 10pm) total mean \\
6 & Morning (6am - 10am) total mean \\
7 & Noon (10am - 2pm) total mean\\
8 & Night (1am - 5am) total mean \\
9 & Week max power \\
10 & Week min power \\
11 & ratio of Mean over Max\\ 
12 & ratio of Min over Mean \\
13 & ratio of Morning over Noon \\
14 & ratio of Noon over Day \\
15 & ratio of Night over Day \\ 
16 & ratio of Weekday over Weekend \\
17 & proportion of time with  $P_t > 0.5 kw$  \\
18 & proportion of time with  $P_t > 1 kw$ \\
19 & proportion of time with  $P_t > 2 kw$ \\
20 & sample variance of $P_t$  \\
21 & sum of difference $|P_t - P_{t-1}|$ \\
22 & sample cross correlation of subsequent days \\ 
23 & number of counts that $|P_t - P_{t-1}| > 0.2 kw$ \\
\hline 
\end{tabular}
\end{table}
%
%%%%%%%%%%%%%%%%%%%%%%%%
%%%%%%%%%%%%%%%%%%%%%%%%%%%%
% list images samples
%
% \vspace*{-0.19in}
\subsection{Images}
\label{cdc:appx:original:sampled:imgs}
\begin{figure}[!hbpt]
\centerline{\includegraphics[width=0.95\columnwidth]{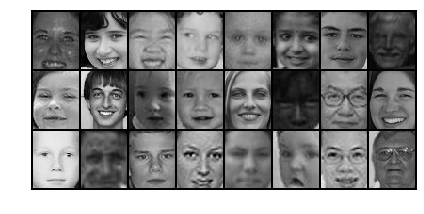}}
% \vspace*{-0.19in}
\caption{The sampled images with the dimension of $61 \times 49$ for each one. }
\label{cdc:appx:original:sampled:group_ppl}
\end{figure}

% \vspace*{-0.19in}
Original people images are shown in Figure~\ref{cdc:appx:original:sampled:group_ppl}
%%%%%%%%%%%%%%%%%%%%%%%%%%
%%%% low rank display %%%%
%%%%%%%%%%%%%%%%%%%%%%%%%%
\subsection{Low rank linear transformation}
\begin{figure*}[!bpht]
\centering
\framebox{\parbox[b]{1.92in}
{\includegraphics[scale=0.26]{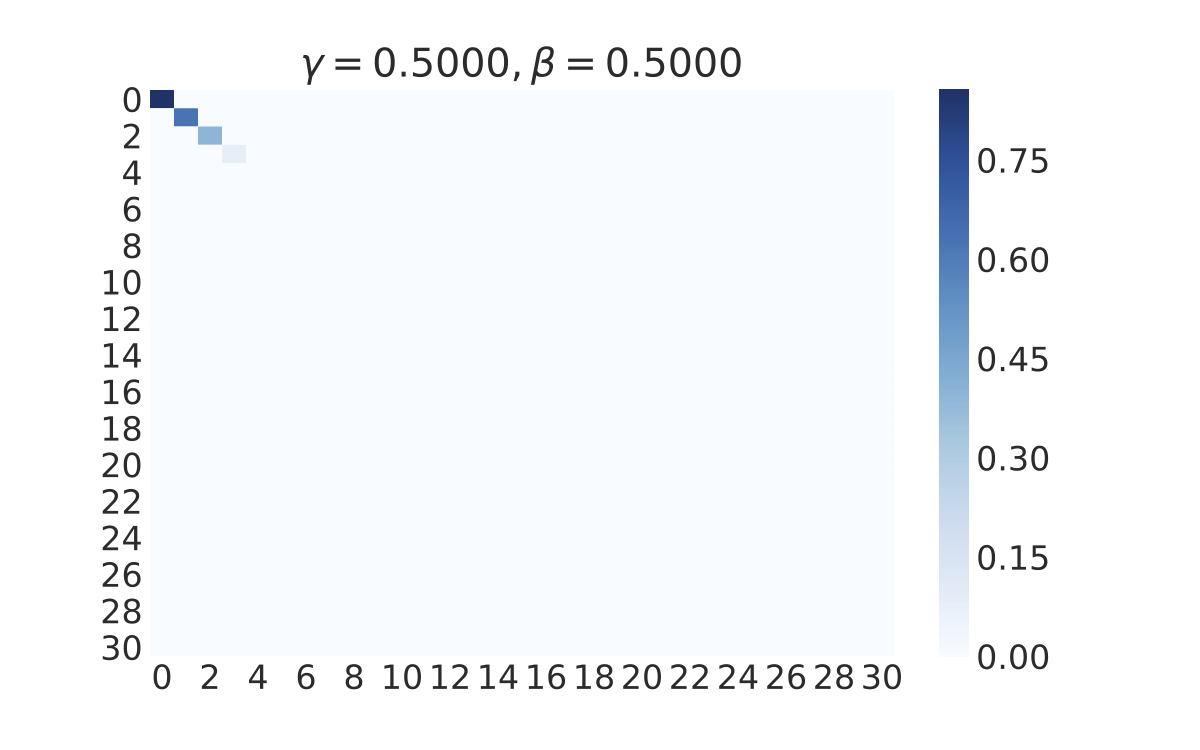}
}
{\includegraphics[scale=0.26]{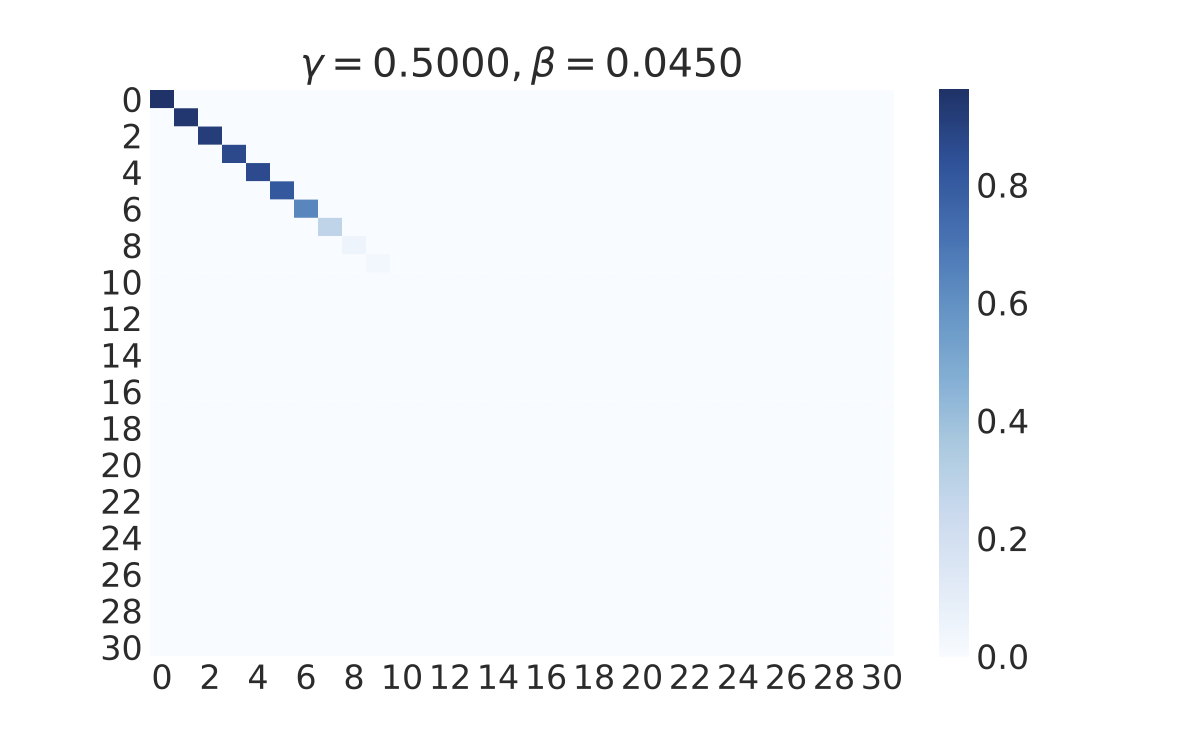}
}
{\includegraphics[scale=0.26]{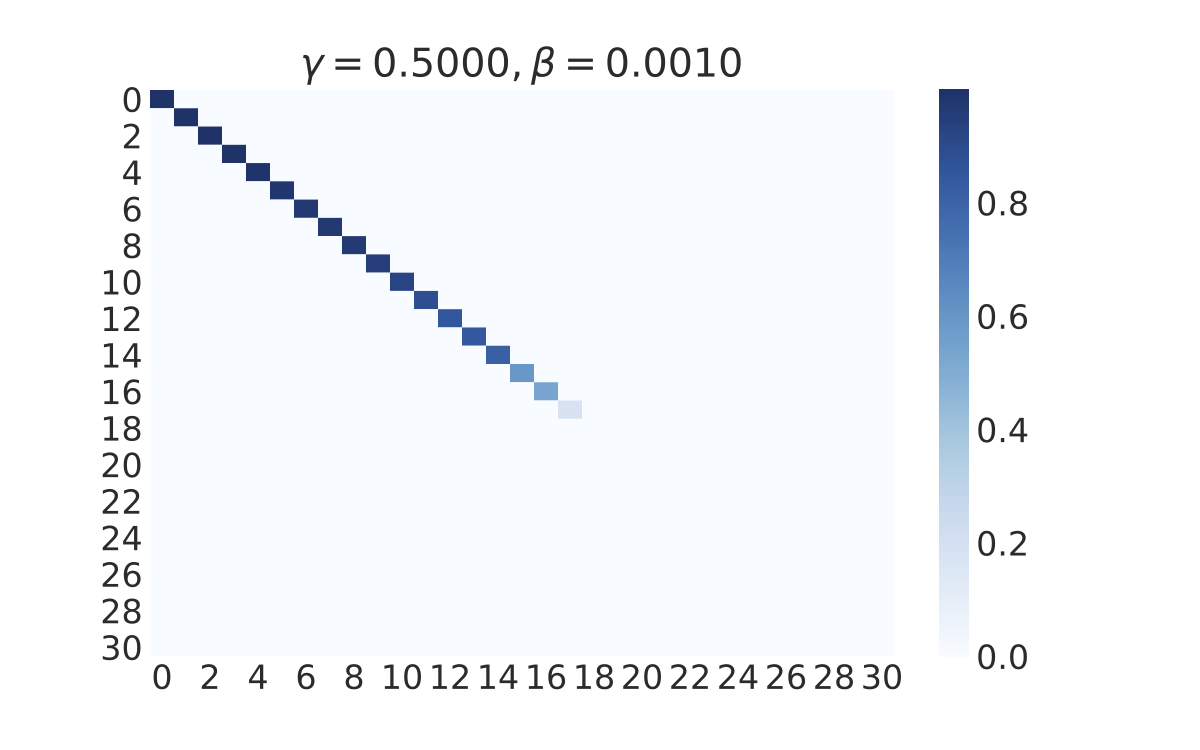}
}}
\caption{Singular values for a sampled data patch when the rank equals 4, 10, and 18}
\label{cdc:linear:singular:val:matrix} 
\end{figure*}

\begin{figure}[!ht]
\vskip -0.12in
% \begin{center}
\framebox{\parbox[b]{1.49in}
{\includegraphics[scale=0.20]{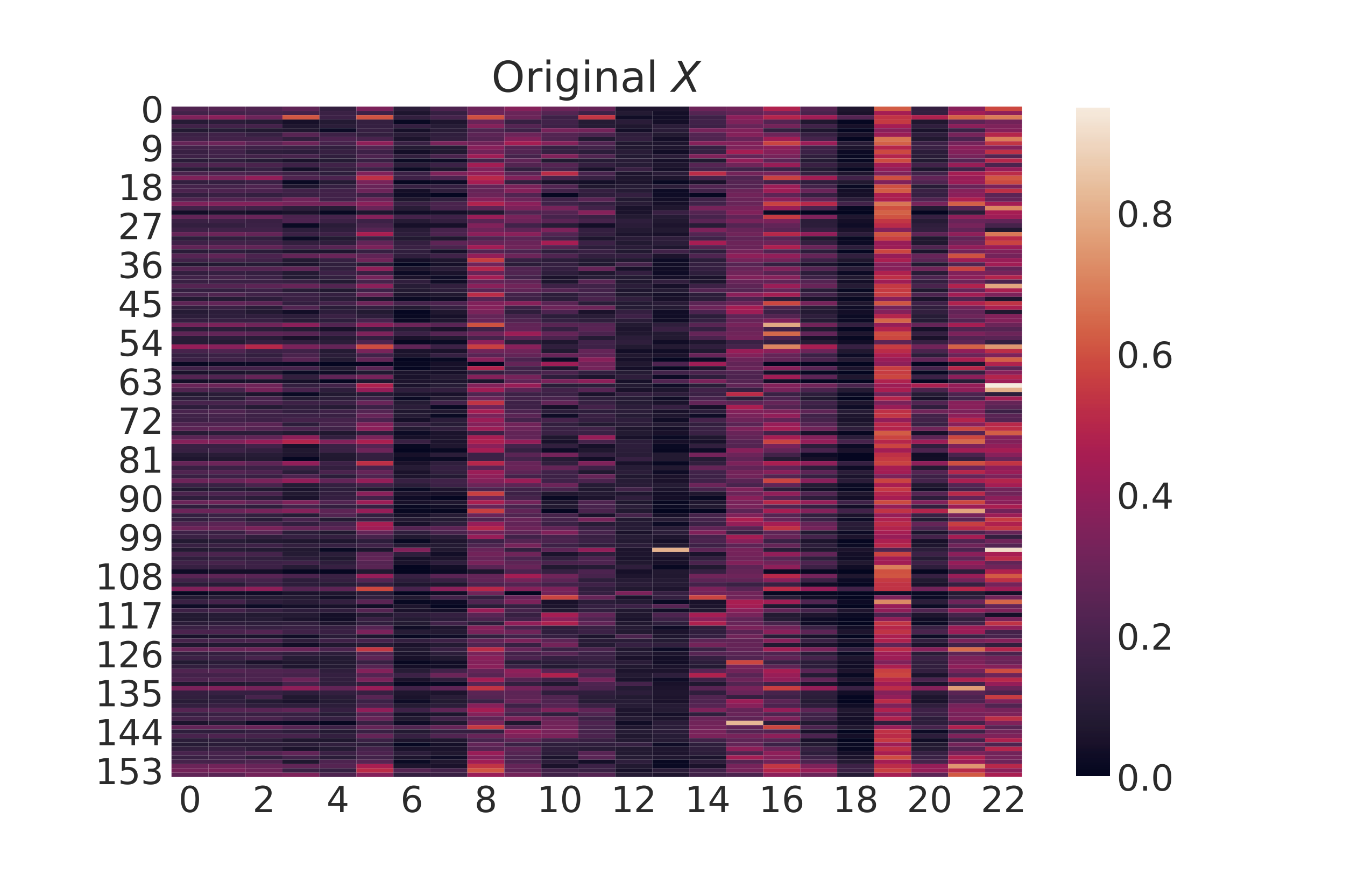}}
\hspace{-0.01in}
{\includegraphics[scale=0.20]{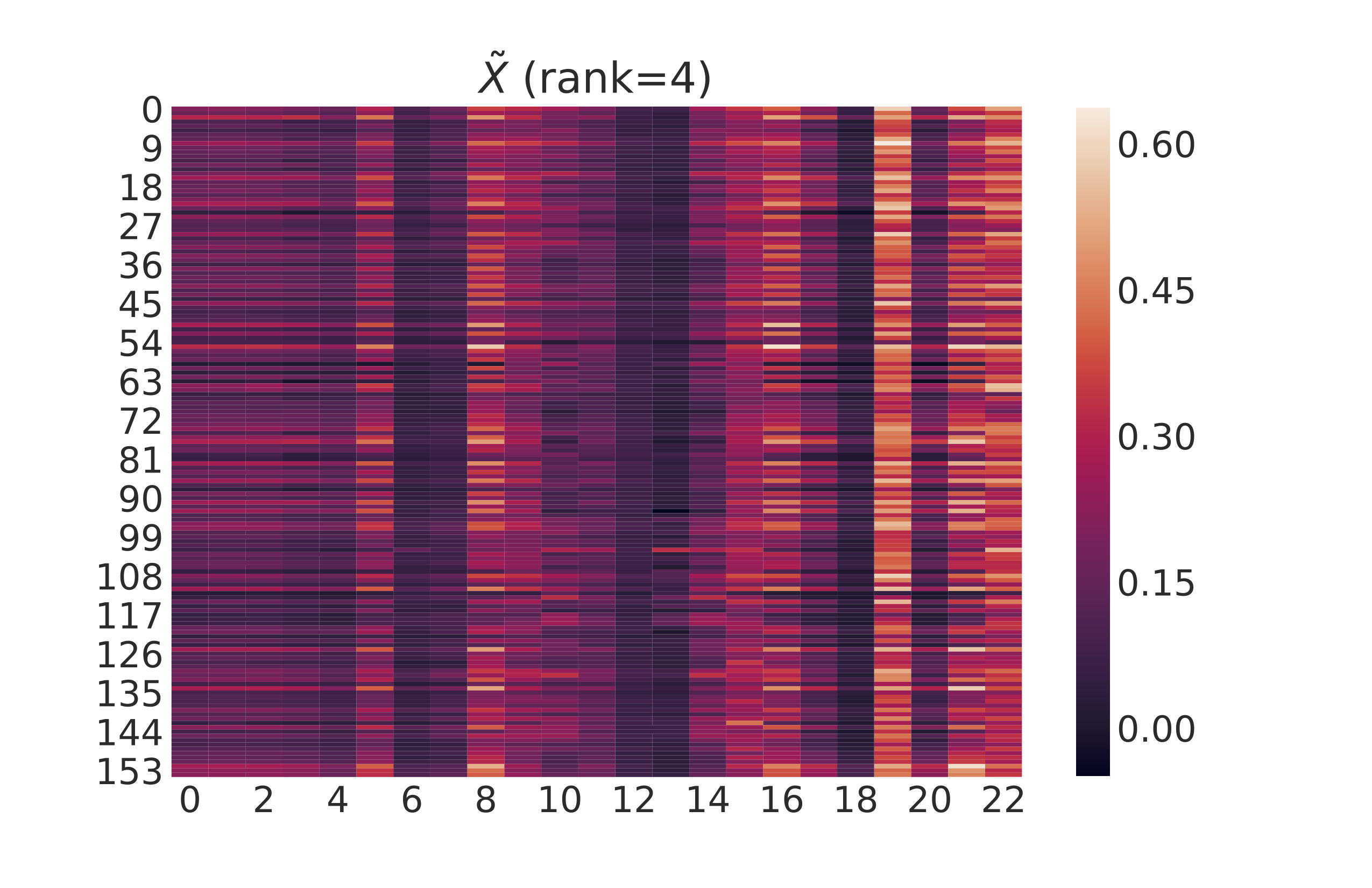}
}}
\framebox{\parbox[b]{1.485in}
{\includegraphics[scale=0.204]{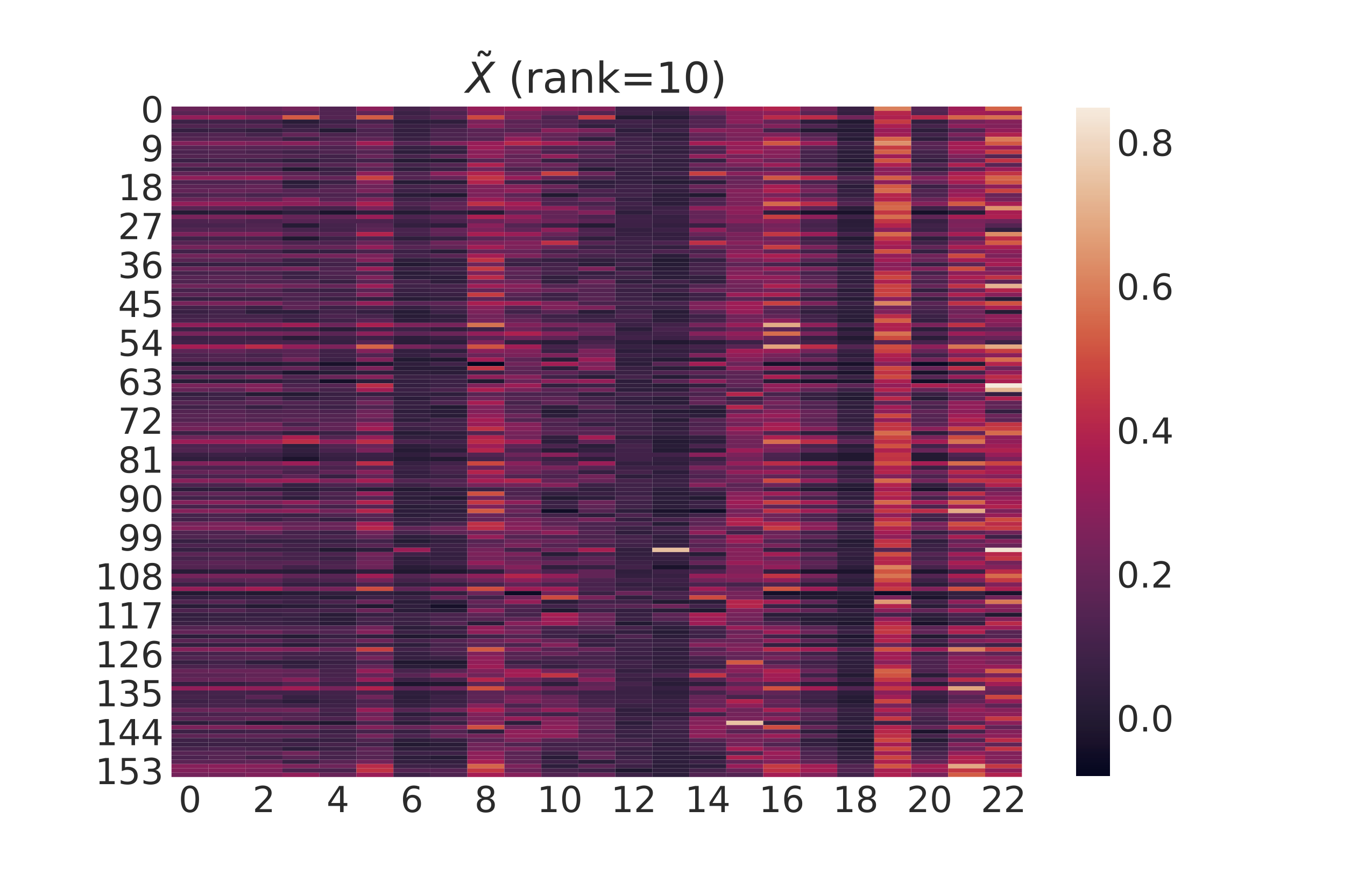}}\hspace{-0.01in}
{\includegraphics[scale=0.204]{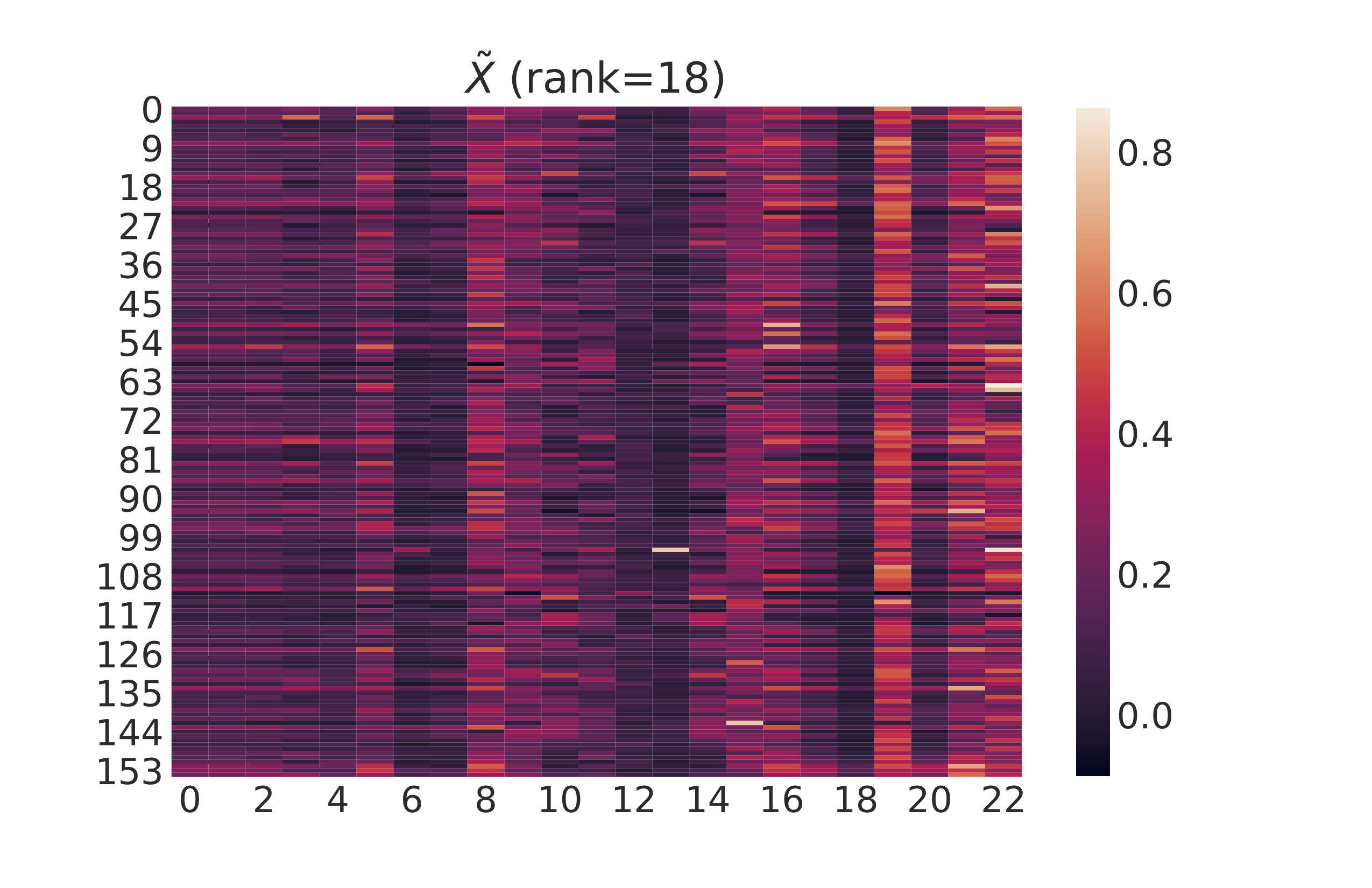}
}}
% \end{center}
\vspace{-0.10in}
\caption{The data holder applies a linear transformation on the data, i.e. compresses it down to low rank matrices and recovers it back to the original dimension. Each figure consists of 155 sampled households with 23 features which has been normalized originally. We fix the distortion tolerance $\gamma$ and tune the nuclear norm coefficient $\beta$ to get the result of various rank scenarios. }
\label{priv:fig:linear:svd:multi_rank_cases:heatmap}
\end{figure}

Singular value matrices for a batch of samples with the low rank transformation are shown in Figure~\ref{cdc:linear:singular:val:matrix}. And the raw and altered power consumption features are shown in Figure~\ref{priv:fig:linear:svd:multi_rank_cases:heatmap}
% %

% \begin{figure*}[!bpht]
% \centering
% % \begin{minipage}{0.99\textwidth}
% \framebox{\parbox[b]{2.0in}
% {\includegraphics[scale=0.26]{fig_linear//margin_orig_alter_rank_4_heatmap.eps}
% }
% {\includegraphics[scale=0.26]{fig_linear//margin_orig_alter_rank_10_heatmap.eps}
% }
% {\includegraphics[scale=0.26]{fig_linear/margin_orig_alter_rank_18_heatmap.eps}
% }}

% \caption{Difference between altered and original data when rank = 4, 10, and 18}
% % \end{minipage}
% \end{figure*}

%---add on the sample visualization---%
% ----------------------------

%
%
% 